\documentclass[]{fairmeta}
\usepackage{lineno}
\usepackage{calc}
\usepackage{amsmath}
\usepackage{caption}
\usepackage{multirow}
\usepackage{amsfonts}
\usepackage{xspace}
\usepackage[font=footnotesize]{subcaption}
\usepackage{booktabs}
\usepackage{pdfpages}
\usepackage{xcolor}
\usepackage{tabularx}
\usepackage{bbm}
\usepackage{graphicx}
\usepackage{algorithm}
\usepackage{algorithmic}
\usepackage{enumitem}
\usepackage{pifont}

\usepackage{microtype}
\usepackage{hyperref}
\usepackage{url}
\usepackage{booktabs}
\usepackage{graphicx}
\usepackage{xspace}
\usepackage{subcaption}
\usepackage{amsmath}
\usepackage{wrapfig}
\usepackage{longtable}
\usepackage{booktabs}
\usepackage{enumitem}
\usepackage{multirow}

\usepackage[export]{adjustbox}

\newcommand{\ours}{\textsc{SandMLE}\xspace}

\newcommand{\tabref}[1]{Table~\ref{#1}}
\newcommand{\figref}[1]{Fig.~\ref{#1}}

\newcommand{\secref}[1]{\S\ref{#1}}
\newcommand{\appref}[1]{Appendix~\ref{#1}}

\title{
Synthetic Sandbox for Training Machine Learning Engineering Agents
}

\author{Yuhang Zhou$^{*}$}
\author{Lizhu Zhang$^{*}$}
\author{Yifan Wu}
\author{Jiayi Liu}
\author{Xiangjun Fan}
\author{Zhuokai Zhao$^{\dagger}$}
\author{Hong Yan$^{\dagger}$}

\affiliation{Meta AI}

\contribution[*]{Equal contribution}
\contribution[\dagger]{Joint last author}

\abstract{
As large language model agents advance beyond software engineering (SWE) tasks toward machine learning engineering (MLE), verifying agent behavior becomes orders of magnitude more expensive: while SWE tasks can be verified via fast-executing unit tests, MLE verification requires running full ML pipelines---data preprocessing, model training, and metric evaluation---on large datasets at each rollout step, rendering trajectory-wise on-policy reinforcement learning (RL) prohibitively slow.
Existing approaches retreat to supervised fine-tuning (SFT) or offline proxy rewards, sacrificing the exploration and generalization benefits of on-policy RL.
We observe that \textit{sandbox data size} is the primary source of this bottleneck.
Based on this insight, we introduce \textbf{\ours}, a multi-agent framework that generates diverse, verifiable synthetic MLE environments from a small number of seed tasks, preserving the structural and technical complexity of real-world problems while constraining datasets to micro-scale (each task is paired with only 50–200 training samples).
Through extensive experiments, we show that \ours reduces execution time by over 13×, enabling large-scale, on-policy trajectory-wise RL for the first time in the MLE domain.
On MLE-bench-lite, \ours yields significant gains over SFT baselines across Qwen3-8B, 14B, and 30B-A3B, with relative medal rate improvements ranging from 20.3\% to 66.9\%.
Furthermore, the trained policy generalizes across unseen agentic scaffolds, achieving up to 32.4\% better HumanRank score on MLE-Dojo.
}

\date{\today}
\correspondence{Yuhang, Lizhu, Zhuokai, Hong at \email{\{zyhang, lizhu, zhuokai, hong\}@meta.com}\\\\}

\begin{document}

\maketitle

\section{Introduction}
\label{sec:intro}

Large Language Models (LLMs) have demonstrated exceptional reasoning capabilities, particularly when optimized through Reinforcement Learning with Verifiable Rewards (RLVR)~\citep{guo2025deepseek}.
However, as these models are deployed to solve increasingly complex problems, machine learning engineering (MLE) has emerged as a challenging frontier.
Unlike simple question answering, MLE tasks inherently cannot be resolved in a single turn; they require an agent to iteratively propose ideas, write code, execute it, analyze environment feedback, and adjust its strategy over multiple attempts.
Most existing work on applying LLM agents to MLE tasks focuses on developing complex agent scaffolding to improve task performance~\citep{toledo2025ai, liu2025ml}, but these framework-level enhancements do not fundamentally improve the agent's intrinsic reasoning abilities.
\begin{figure}[t]
    \centering
    \includegraphics[width=\linewidth]{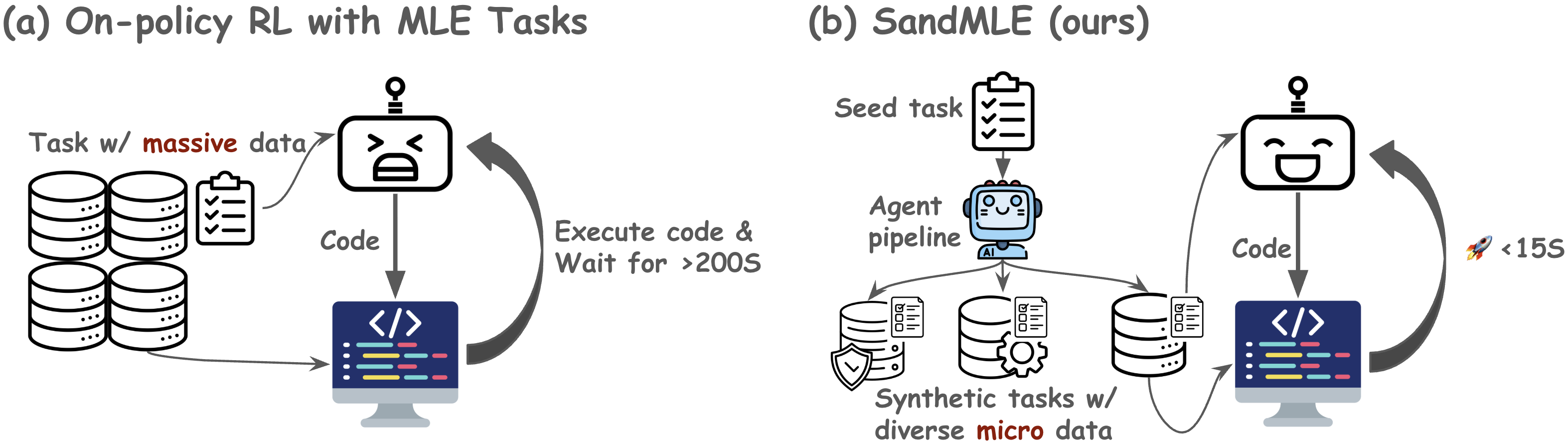}
    \vspace{-0.1in}
    \caption{
        Standard on-policy RL for MLE tasks (left) is bottlenecked by execution latency: each rollout step requires running full ML pipeline on large associated datasets ($>$200s).
        \ours transforms seed tasks into diverse synthetic environments with micro-scale datasets ($<$15s), making trajectory-wise on-policy RL practically feasible.
        %
        %
        }
    \label{fig:teaser}
    \vspace{-0.1in}
\end{figure}

For similar long-horizon challenges, such as software engineering (SWE)~\citep{jimenez2024swebench} and web search~\citep{zhou2023webarena}, previous works have successfully utilized trajectory-wise reinforcement learning (RL) to enhance LLM reasoning~\citep{wu2025webdancer, deepswe2025, wu2025resum, chen2025scaling}.
This training paradigm extends rollouts across multiple turns, assigning rewards based on the final outcome of the trajectory to optimize the model's long-horizon decision-making capabilities.
In SWE tasks, unit tests execute in seconds; while in MLE, a single code execution may require training a model from scratch.
This orders-of-magnitude gap in feedback latency makes applying online RL to MLE a critical bottleneck: each rollout step requires executing full ML pipelines---data preprocessing, model training, and metric evaluation---with a single code execution averaging nearly 200 seconds on standard problems.
%
%
This computational overhead renders the large-scale rollouts required for trajectory-wise RLVR highly impractical.
%
%
Consequently, the few existing efforts that do attempt to train models for these MLE tasks have largely retreated to supervised fine-tuning (SFT) or step-wise RLVR using proxy offline rewards~\citep{cai2026acegrpo, liu2025mlagent}, sacrificing the exploration and generalization benefits typically achieved through on-policy algorithms like Group Relative Policy Optimization (GRPO)~\citep{shao2024deepseekmath} and others~\citep{chu2025sft, zhou2025disco, yu2025dapo, yang2025let}.

A closer analysis reveals a surprisingly simple root cause.
Unlike SWE, where execution time is dominated by compilation and test logic, MLE latency is overwhelmingly driven by dataset size.
%
%
Based on this insight, we introduce \ours (illustrated in Figure \ref{fig:teaser}), a novel paradigm that utilizes multi-agent collaboration to generate synthetic MLE sandboxes where both training and test data are intentionally constrained to micro-scale (50-200 samples), reducing average execution time to under 15 seconds and enabling thousands of on-policy rollout updates where previously only a handful were feasible.
%
For the data construction pipeline of \ours, we design a fully-automated, agent-driven framework that generates diverse, verifiable synthetic MLE tasks.
By drastically reducing the data volume while preserving the structural complexity and mathematical hidden rules of the machine learning (ML) problem, \ours facilitates high-speed environment feedback, making large-scale, online rollouts practical.

Specifically, our pipeline orchestrates multiple specialized language model roles (as illustrated in \figref{fig:syntho_diagram}): 
a \textit{Data Strategist} abstracts the structural DNA from seed tasks to formulate novel scenarios; 
an \textit{MLE Developer} procedurally generates the raw data assets and embeds the mathematical hidden rules;
an \textit{MLOps Engineer} constructs deterministic evaluation sandboxes with established performance thresholds;
and a \textit{Technical Writer} synthesizes the final task specifications.
Using this generated synthetic data, we apply a specifically designed reward formulation and training pipeline to execute trajectory-wise RL, directly enhancing the model's MLE problem solving capabilities.

Our main contributions are summarized as follows:
\begin{itemize}[leftmargin=*]
    \item \textbf{Scalable Synthetic MLE Sandbox Generation:} 
    We design a novel multi-agent framework that procedurally extracts structural DNA from seed tasks to generate diverse, high quality, and verifiable synthetic MLE tasks with micro-scale datasets.

    \item \textbf{Trajectory-wise RL for MLE:}
    Leveraging these synthetic environments, we propose and successfully apply trajectory-wise RL with the milestone-based reward to the MLE domain, enabling LLMs to learn complex, long horizon trial-and-error strategies.

    \item \textbf{Framework-Agnostic Generalization:} 
    Through extensive experiments, we demonstrate that models trained via our approach not only achieve significant performance gains, including a 20.3\% to 66.9\% relative improvement in \texttt{Any Medal} rate over the SFT baseline across the Qwen3-8B, 14B, and 30B-A3B models on MLE-bench-lite~\citep{chan2024mle}, but also generate robustly across unseen agentic scaffolds,
    achieving up to 32.4\% relative improvement in HumanRank score on the MLE-Dojo benchmark~\citep{qiang2025mle}.
    %
\end{itemize}
\section{Related Work}
\label{sec:related}

\subsection{Machine Learning Engineering (MLE) for Agents}

The rapid advancement in the reasoning capabilities of LLMs has recently enabled their application to complex, open-ended MLE tasks.
To systematically evaluate these capabilities, MLE-bench~\citep{chan2024mle} was introduced as a major benchmark curating real-world Kaggle competitions with standardized medal-rate evaluation.
Building on this foundation, MLE-Dojo~\citep{qiang2025mle} and MLE-Smith~\citep{qiang2025mles} have expanded the volume and diversity of available MLE tasks to facilitate broader evaluation.

Concurrently, substantial research has focused on enhancing agent performance through sophisticated scaffoldings and test-time compute scaling~\citep{nam2025mle}.
Evolutionary and iterative frameworks, including AIDE~\citep{jiang2025aide}, AIRA~\citep{toledo2025ai}, ML-Master~\citep{liu2025ml, zhu2026toward}, as well as broader systems like the R\&D Agent~\citep{yang2025r} and FM Agent~\citep{li2025fm}, leverage multi-turn environmental feedback to iteratively refine their reasoning trajectories at inference time. 
While these approaches effectively improve task performance, the gains are contingent on specific scaffold design rather than the underlying model's reasoning capacity.
%
Indeed, our own cross-framework evaluation (\tabref{tab:generalization_lite},~\ref{tab:generalization_dojo}) reveals that the same base model can exhibit dramatically different performance depending on the scaffold, underscoring that scaffolding alone does not yield robust, transferable MLE capabilities.
In contrast to these inference-only approaches, \ours targets this gap directly by enabling scalable on-policy RL training, enhancing the model's intrinsic engineering reasoning in a framework-agnostic manner.
%

\subsection{Reinforcement Learning for MLE Agents}
\label{subsec:rl_mle_agents}

Recent works~\citep{zhang2025landscape, singh2025agentic} have demonstrated that execution-based, trajectory-wise RL can effectively optimize agents for long-horizon tasks.
DeepSWE~\citep{deepswe2025} and WebDancer~\citep{wu2025webdancer} show strong results in software engineering and web search respectively, while~\citet{chen2025scaling} demonstrate that synthesizing diverse experience trajectories further improves policy learning.
%
%
However, extending these frameworks to MLE introduces a severe computational bottleneck: each rollout step requires training and evaluating ML models to derive environmental rewards renders standard on-policy RL prohibitively slow~\citep{cai2026acegrpo}.

To bypass this latency, prior approaches have relied on SFT over expert trajectories or utilize asynchronous step-wise RL architectures to hide execution time~\citep{liu2025ml, yang2025reinforcement}. 
Yet, both strategies are fundamentally off-policy; for instance, asynchronous GRPO relies on trajectories generated from lagging policy states, creating a distribution shift~\citep{cai2026acegrpo}. 
In contrast, \ours leverages fast-executing synthetic environments to maintain a strictly on-policy training regime. 
By eliminating the execution bottleneck, we ensure that the accumulated reasoning trajectories are natively generated from the current model state, providing a high-fidelity gradient signal for effective and stable learning.

\section{Preliminaries}\label{subsec:preliminaries}

%

\subsection{MLE as a Sequential Decision Process}\label{subsec:SDP}
We formalize the MLE task as a sequential decision process and review GRPO~\citep{shao2024deepseekmath}, the RLVR algorithm we employ, along with the specific challenges that arise when applying it to multi-turn agentic settings.
Specifically, an MLE task can be defined by a tuple $(I, \mathcal{T}, \mathcal{E})$, where $I$ denotes the initial task specification (dataset description and objective), $\mathcal{T}$ represents the available tool set (e.g., code execution, file I/O, and scoring functions), and $\mathcal{E}$ is the interactive execution environment.

We cast the agent's interaction with $\mathcal{E}$ as a finite-horizon sequential decision problem.
At each step $t$, the language model policy $\pi_\theta$ observes the current trajectory history $h_t = \{I, \mathcal{T}, a_1, o_1, \dots, a_{t-1}, o_{t-1}\}$, comprising the task specification and all prior action–observation pairs. 
Conditioned on $h_t$, the language model policy $\pi_\theta$ generates an action $a_t \sim \pi_\theta(\cdot \mid h_t)$, which typically consists of generating or updating code to construct the ML pipeline.
The environment $\mathcal{E}$ then executes the action $a_t$ and returns an observation $o_t$, which may include standard output, runtime errors, or intermediate evaluation metrics, extending the history to $h_{t+1} = h_t \cup \{a_t, o_t\}$. 
%
%
This iterative loop continues until a termination condition is reached, e.g., the agent emits a designated submission action, or a maximum step limit $T_{\text{max}}$ is reached. 
Upon completion, the agent's final output is evaluated against a hidden test set to yield a scalar score $S$; for example, in the Kaggle-style benchmarks~\citep{chan2024mle}, $S$ is derived from the agent's relative ranking on the task leaderboard.

\paragraph{The Execution-Latency Bottleneck.}
A fundamental bottleneck in this formulation is the environment execution latency.
Unlike unit tests in SWE tasks, MLE tasks inherently require training ML models and performing inference over large datasets, making each environment step orders of magnitude more expensive.
%
%
Because on-policy RL invokes $\mathcal{E}$ at every step of every rollout for every sample in a group, this latency renders optimization prohibitively expensive.
Prior work has therefore typically permitted extended time windows of up to 24 hours for an agent to complete a single task~\citep{toledo2025ai, liu2025ml}.

\subsection{Reinforcement Learning from Verifiable Rewards (RLVR)}\label{subsec:rlvr}

RLVR leverages objective, rule based feedback to optimize LLMs.
GRPO~\citep{shao2024deepseekmath} is an efficient variant of Proximal Policy Optimization~\citep{schulman2017proximal} that eliminates the need for an independent value network or critic model.
Instead, GRPO evaluates the relative quality of responses within a specifically sampled group.
For a given input query $x$, the old policy $\pi_{\text{old}}$ samples a group of $N$ candidate outputs $\{y_1, y_2, \dots, y_N\}$.
Each candidate output receives a verifiable reward $r_i$ from the environment.
The advantage $\hat{A}_i$ for each specific output is computed by normalizing the rewards strictly within the sampled group: 
$ \hat{A}_i = \frac{r_i - \mu_r}{\sigma_r}, $
where $\mu_r$ and $\sigma_r$ are the mean and standard deviation of the group rewards.
The GRPO objective maximizes this relative advantage while simultaneously penalizing the KL divergence from a reference model $\pi_{\text{ref}}$ to prevent policy collapse:
\begin{equation}
    J_{\text{GRPO}}(\theta) = \mathbb{E}_{x \sim P, y_i \sim \pi_{\text{old}}} \left[ \frac{1}{N} \sum_{i=1}^N \min \left( \rho_i \hat{A}_i, \text{clip}(\rho_i, 1-\epsilon, 1+\epsilon) \hat{A}_i \right) - \beta \mathbb{D}_{\text{KL}}(\pi_\theta \| \pi_{\text{ref}}) \right]
    \nonumber
\end{equation}
where $\rho_i = \frac{\pi_\theta(y_i|x)}{\pi_{\text{old}}(y_i|x)}$ is the probability ratio, $\epsilon$ is the surrogate clipping threshold, and $\beta$ is the coefficient controlling the KL penalty.
%

\paragraph{Applying GRPO to Multi-turn, Trajectory-level Agentic Tasks.}
GRPO is agnostic to the internal structure of each output $y_i$, i.e., the algorithm itself is unchanged whether $y_i$ is a single-turn response or a multi-turn trajectory.
In the agentic MLE setting, each $y_i$ is no longer a single model generation but a trajectory $\tau_i = \{a_1, o_1, \dots, a_T, o_T\}$ in which model-generated actions $a_t$ alternate with environment-generated observations $o_t$, and the policy gradient is computed only over action tokens while masking observations from the loss~\citep{deepswe2025, wu2025webdancer}.
This multi-turn structure introduces two practical challenges that make naive application difficult.
First, \textit{reward sparsity}: unlike single-turn RLVR (e.g., math or QA), where reward is available immediately after generation, the agentic reward $r_i$ is determined only upon trajectory completion, after potentially dozens of action-observation exchanges, making credit assignment substantially harder. 
Second, \textit{execution cost}: each rollout step invokes the environment $\mathcal{E}$, and each trajectory is sampled $N$ times per input for group normalization, so the total wall-clock cost scales as $O(N \cdot T_{max} \cdot c_{exec})$ per training example.
When $c_\text{exec}$ is on the order of minutes, as in standard MLE benchmarks, on-policy training becomes infeasible.
Our approach addresses both two challenges directly: \ours reduces $c_\text{exec}$ by over $13\times$ through synthetic micro-scale environments (\secref{subsec:synthetic_pipeline}), and we design a dense, milestone-based reward to mitigate sparsity (\secref{subsec:trajectory_grpo}).

%
%
%
%
\section{\ours}\label{sec:workflow}
To enable computationally feasible trajectory-wise RL for MLE, we must first overcome the prohibitive execution latency of real-world tasks.
While downsampling datasets seems like a natural solution, it fundamentally corrupts the evaluation environment by breaking fixed test sets and invalidating established leaderboard baselines.
Furthermore, downsampling alone does not address the scarcity of diverse training tasks required for robust policy optimization.
To address these limitations, we propose \ours, a framework that algorithmically generates a massive, diverse curriculum of synthetic MLE tasks with their associated data and environment.
By constructing these sandboxes from the ground up at a micro-scale, we ensure that the entire pipeline---from preprocessing to inference---executes in a fraction of the time. 
This strategic reduction preserves the integrity of the reward signal while scaling the number of unique training environments, transforming GRPO into a feasible on-policy training paradigm for MLE agents.
\begin{figure*}[t]
    \centering
    \includegraphics[width=\linewidth]{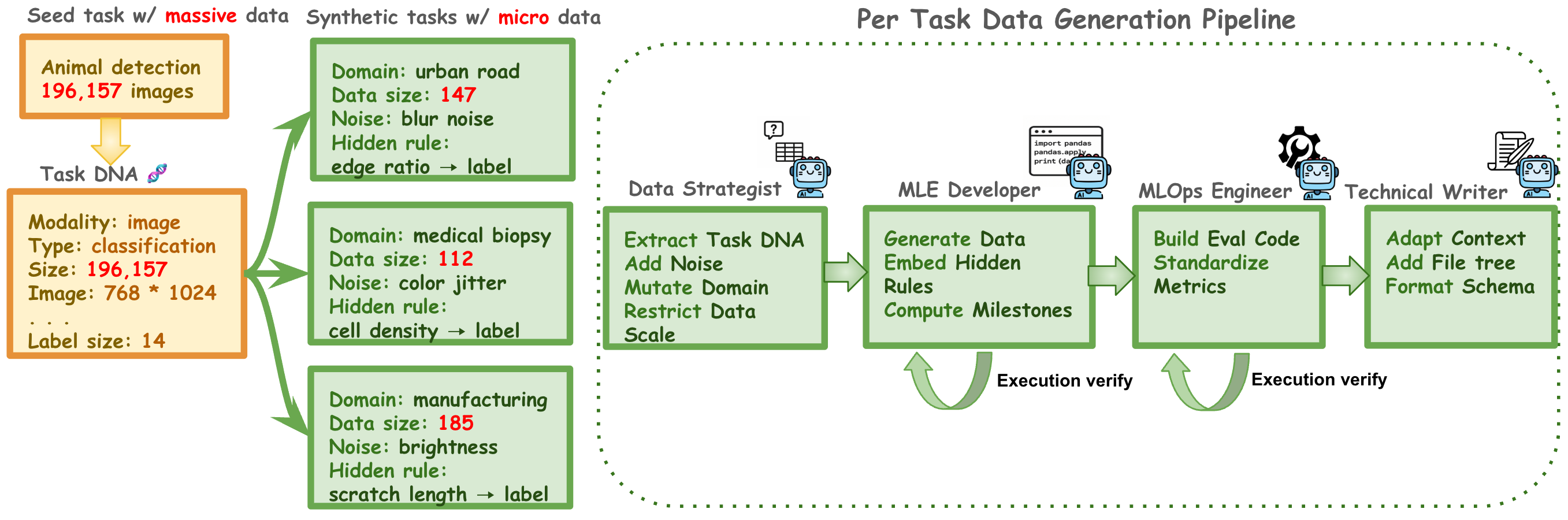}
    \caption{
        The Agentic MLE Environment Factory.
        %
        Illustrated is the procedural multi-agent workflow that transforms a massive, slow-executing seed task into a coherent, high-speed, verifiable \textit{synthetic micro-task}.
        The pipeline explicitly optimizes for data efficiency by strictly controlling the task-associated training data size (down from 196,157 to less than 200) to enable rapid iteration during policy optimization.
        %
    }
    \label{fig:syntho_diagram}
    \vspace{-0.2in}
\end{figure*}

\subsection{Synthetic Environment Generation}\label{subsec:synthetic_pipeline}
\paragraph{Pipeline Overview.}
As illustrated in \figref{fig:syntho_diagram}, our automated road map transforms curated seed tasks into lightweight synthetic environments through four interconnected phases:
1. \textit{Task Amplification and Specification}: An agent extracts the structural blueprint of a real-world seed task, alters its application domain, and defines underlying mathematical rules and noise distributions;
2. \textit{Agentic Data Generation}: A code-driven approach is employed to procedurally generate the small-scale datasets, applying the hidden rules to establish ground truth labels while implementing baseline models to verify learnability; 
3. \textit{Evaluation Environment Setup}: We construct an automated sandbox to safely execute the policy's generated code and calculate final metrics against a hidden test set; 
4. \textit{Task Description Synthesis}: Finally, the original seed task's narrative is adapted to precisely match the newly generated domain, feature schema, and evaluation metrics, ensuring an accurate and cohesive prompt for policy optimization.
We illustrate each components with details next.

\paragraph{Data Strategist: Seed Task Amplification and Specification.}
\label{subsec:task_amplification}
To maximize the diversity of our synthetic environments, a \textit{Data Strategist} agent extracts and manipulates the core structure of real-world seed tasks through four steps: 
(i) \textit{Structural DNA Extraction}: the agent abstracts a seed task into a \textit{Task DNA}, which is a mathematical schema (e.g., modality, resolution, label cardinality) stripped of semantic context.
(ii) \textit{Domain Attribution}: the abstracted DNA is then mapped to broader scenarios (e.g., re-purposing an animal image classification into road damage detection).
(iii) \textit{Adversarial Mutation}: the agent injects realistic difficulty via a noise configuration $\epsilon$ (e.g., image blurring).
%
%
And (iv) \textit{Concrete Specification Compilation}: the agent merges these elements, defining a complex hidden rule $H: l = f(z) + \epsilon$ that connects individual features $z$ to labels $l$ in the synthetic micro dataset $Z$.
For instance, in an urban road damage detection task, the rule might dictate that a specific edge pixel ratio ($f(z)$) combined with injected blur noise ($\epsilon$) strictly determines the final damage severity label ($l$). 
The agent simultaneously restricts the total dataset size to 50--200 samples to guarantee rapid execution latency later during MLE training.

To mitigate the challenge of sparse reward signals in later policy optimization, the agent also specifies multiple baseline methods of varying complexity. 
These methods theoretically establish a set of progressive milestone thresholds $\mathcal{S}$ to enable dense reward assignment.

\paragraph{ML Developer: Synthetic Data Generation.}
\label{subsec:agentic_data_generation}
To translate these abstract specifications into executable environments, an \textit{ML Developer} agent writes a self-contained Python script to perform four critical operations.
First, it synthesizes the micro-scale datasets $Z$ and partitions them into training ($\mathcal{D}_{\text{train}}$) and hidden test ($\mathcal{D}_{\text{test}}$) sets. 
Then, it implements the defined hidden rules $H$ to deterministically map the generated features $z$ to ground-truth labels $l$ via $H: l = f(z) + \epsilon$.
Continuing our previous example of urban road damage detection task, the agent writes the procedural code to generate the synthetic urban road images, calculates the edge pixel ratio $f(z)$ for each, injects the specified blur noise $\epsilon$, and mathematically assigns the final damage severity label $l$.
%
Next, the script trains and evaluates the \textit{Data Strategist}'s specified baseline methods to empirically calculate and lock in the set of progressive milestone thresholds $\mathcal{S}$ required for downstream dense reward calculation.
Finally, it outputs a dummy sample submission file containing the test data paired with random predictions to serve as a strict schema reference. 
To ensure environment reliability, we enforce an \textit{execution-based verification loop}: if the script fails, the current trace is automatically returned to the agent for iterative debugging.
%

\paragraph{MLOps Engineer: Evaluation Environment Setup.}
\label{subsec:eval_environment}
To construct the automated scoring sandbox, an \textit{MLOps Engineer} agent writes a robust \texttt{evaluator script} based on the task specification and training data $\mathcal{D}_{\text{train}}$. 
This process involves three key requirements: 
First, \textit{metric standardization}: the agent hard-codes the evaluation metric $\mathcal{M}$, the optimization direction, and the set of progressive milestone thresholds $\mathcal{S}$ to ensure deterministic scoring.
Second, \textit{data alignment and computation}: the script loads an agent's submission (containing predictions $\hat{p}$), aligns it with the hidden ground truth labels $l \in \mathcal{D}_{\text{test}}$, computes the exact metric $\mathcal{M}(\hat{p}, l)$, and outputs the final score. 
And third, \textit{execution-based verification}: to guarantee reliability, the system automatically tests the \texttt{evaluator script} against a dummy sample submission.
%
Any runtime errors are fed back to the \textit{MLOps Engineer} for iterative debugging to finalize a functional evaluation environment.

\paragraph{Technical Writer: Task Description Synthesis.}
\label{subsec:task_description}
Once the data and environments are fully generated, a \textit{Technical Writer} agent compiles the metadata into a comprehensive markdown document. 
This synthesis involves four steps: 
(i) \textit{contextual integration}: the agent adapts the narrative structure of the original seed task to seamlessly match the new synthetic domain, updating names and problem contexts. 
(ii) \textit{content generation}: the agent then drafts a clear problem overview, with specific data format and required evaluation metric. 
(iii) \textit{submission formatting}: the agent explicitly defines the expected output format using the generated sample submission schema. 
and (iv) \textit{file transparency}: finally, the agent catalogs all public data files (e.g., training data, test features) while strictly omitting hidden answers to prevent leakage. 
This documentation pipeline ensures the final synthetic task prompt, which serves as the initial task specification $\mathcal{I}$ as in \secref{subsec:SDP}, is realistic and perfectly aligned with the evaluation code. 
The prompt templates for all four agents are detailed in \appref{sec:prompt}.

\subsection{Environment Sanity Verification}
\label{sec:sanity}

To guarantee the logical consistency of our generated environments prior to RL, we perform an automated sanity check on the evaluation metrics and the established milestone thresholds.
For each generated task, the evaluation script computes a baseline score $s_{\text{sample}}$ derived from a dummy sample submission
, alongside the set of predefined, progressive milestone thresholds $\mathcal{S} = \{s_1, s_2, \dots, s_k\}$, where $s_1$ represents the most rigorous performance standard and $s_k$ represents the most basic baseline.
We set a strict mathematical ordering of these thresholds based on the optimization direction of the task's specific metric $\mathcal{M}$.
Let the boolean indicator $I \in \{0, 1\}$ denote whether a lower score is better for metric $\mathcal{M}$.
A synthetic task is successfully verified and retained only if its thresholds satisfy the following strict monotonic constraint:
\begin{equation}
    \begin{cases}
        s_1 < s_2 < \dots < s_k \land s_1 < s_{\text{sample}}, & \text{if } I = 1 \\
        s_1 > s_2 > \dots > s_k \land s_1 > s_{\text{sample}}, & \text{if } I = 0
    \end{cases}
\end{equation}
Tasks failing to meet this logical ordering are identified as corrupted and are automatically discarded from the training curriculum, ensuring the RL agent only optimizes against valid, monotonic reward signals.

\subsection{Trajectory-Level GRPO}\label{subsec:trajectory_grpo}
\paragraph{Enable On-policy RL for MLE Tasks.}
While trajectory-level GRPO has proven effective for SWE and web search tasks~\citep{deepswe2025, wu2025webdancer}, applying it to MLE tasks has remained computationally infeasible.
The synthetic micro-scale environments constructed by our pipeline (\secref{subsec:synthetic_pipeline}) reduce the wall-clock cost and transforms trajectory-level GRPO from a theoretical possibility into a practical training paradigm for MLE agents.

Concretely, during the rollout phase, the LLM agent interacts with a \ours environment using the ReAct framework~\citep{yao2022react}.
Starting from the initial task specification $\mathcal{I}$, the agent iteratively generates actions (code or reasoning) and receives observations (execution output, errors, or intermediate metrics) over multiple turns using the available tool set $\mathcal{T}$.
To ensure computational efficiency and prevent infinite generation loops, we enforce strict boundary conditions on the environment: a per-step execution time limit $\tau_{\max}$ and a maximum trajectory length $T_{\max}$. 
The agent's trajectory terminates when it yields a final submission, reaches $T_{\max}$, or encounters a terminal error.

\paragraph{Dense Reward Formulation.}
In complex agentic tasks, relying solely on the final performance metric often results in a sparse reward signal, making policy optimization highly unstable. 
To mitigate this sparsity, we design a dense reward function composed of two primary components: a \textit{format reward} and a \textit{milestone-based reward}. 
The final verifiable reward $r$ provides granular feedback across the entire trajectory and is defined as follows:
\begin{equation}
    r = w_{\text{format}} \cdot r_{\text{format}} + w_{\text{execute}} \cdot \mathbb{I}_{\text{execute}} + \sum_{i=1}^{k} w_{s_i} \cdot \mathbb{I}_{s_i}
    \nonumber
\end{equation}
Here, the format reward $r_{\text{format}} \in [0, 1]$ represents the ratio of generated steps that properly utilize the required \texttt{<think>...</think>} reasoning tags. 
The milestone-based component evaluates the final state of the environment using boolean indicator functions ($\mathbb{I} \in \{0, 1\}$). 
This includes basic execution milestones---such as successfully generating and formatting a valid output file ($\mathbb{I}_{\text{execute}}$)---as well as tiered performance milestones ($\mathbb{I}_{s_i}$) that evaluate whether the final quantitative score surpasses the predefined thresholds in our progressive set $\mathcal{S}$. 
By distributing fixed weights ($w$) across these terms to sum to a maximum of $1.0$, this milestone-based reward smoothly guides the agent from basic formatting compliance and code execution to state-of-the-art performance.
%

\paragraph{Selective Masking for Backpropagation.} 
Following previous work for trajectory-level GRPO, we apply strict loss masking during backpropagation to calculate the policy gradient exclusively on the agent's generated reasoning and action tokens, and we entirely mask trajectories where code execution exceeds the defined time limit, thereby preventing the incorrect optimization of static environmental observations and prompts.
\section{Experiments}\label{sec:experiments}

\subsection{Experimental Setup}
\paragraph{Dataset.}
We construct our training corpus \textit{seeded} from the MLE-bench dataset~\citep{chan2024mle}.
Specifically, we collect 60 questions spanning the \texttt{Medium}, \texttt{Hard}, and \texttt{Dev} splits as seeds to ensure exposure to diverse reasoning patterns.
These seeds undergo domain mutation and difficulty boosting, followed by automated sanity check to filter for resolvability and schema validity, as discussed in \secref{sec:workflow}. 
This process yields a final training dataset of 848 synthetic tasks for training and 64 held-out synthetic tasks for validation. 
To evaluate real-world generalization, we employ two separated benchmarks: MLE-Bench-Lite, comprising 22 unseen questions from the \texttt{Easy} split, and MLE-Dojo~\citep{qiang2025mle}, a curated collection of 62 additional tasks derived from broader Kaggle competitions.

\paragraph{Evaluation Metrics.}
Following the evaluation protocol established by MLE-Bench~\citep{chan2024mle}, we measure performance using a hierarchy of success metrics: \texttt{Valid Submission}, \texttt{Above Median}, \texttt{Bronze}, \texttt{Silver}, \texttt{Gold}, and \texttt{Any Medal}. 
These Kaggle-style tiers serves as the concrete instantiation of the abstract progressive milestone set $\mathcal{S} = \{s_1, s_2, \dots, s_k\} = \{s_\text{median}, s_\text{bronze}, s_\text{silver}, s_\text{gold}\}$ introduced in \secref{sec:sanity}, with each tier corresponding to a progressively more demanding leaderboard percentile.
Among these, \texttt{Any Medal} (the union of \texttt{Bronze}, \texttt{Silver}, and \texttt{Gold}) serves as our primary metric.
For MLE-Dojo~\citep{qiang2025mle}, we adhere to the original setup by reporting the \texttt{Valid Submission} rate and the \texttt{Human Rank Score}, which normalizes agent performance against human participants.
More detailed definitions for all metrics are provided in \appref{sec:metric_definition_details}.

\paragraph{Agent Scaffolds.}
Recognizing that model performance is heavily influenced by the choice of agentic scaffolding, we evaluate both baseline models and models finetuned with \ours across a diverse set of agent frameworks to demonstrate their generalization capabilities.
%
For MLE-Bench-Lite, we employ ReAct (consistent with our GRPO rollout)~\citep{yao2022react}, AIRA~\citep{toledo2025ai}, and AIDE~\citep{jiang2025aide}. 
For MLE-Dojo benchmark, we employ its native MLE Agent scaffold~\citep{qiang2025mle} and AIDE, ensuring a robust assessment of our model's adaptability to different agent architectures.

\paragraph{Models and Baselines.}
We apply the proposed \ours training pipeline to three LLMs with varying sizes from the Qwen3 family~\citep{yang2025qwen3}: \texttt{Qwen3-8B}, \texttt{Qwen3-14B} and \texttt{Qwen3-30B-A3B-2507}, and compare the resulting models against the baselines. 
Specifically, \texttt{Base} refers to the off-the-shelf models.
\texttt{Seed-SFT} is a supervised finetuning baseline designed to disentangle the benefit of high-quality seed data from RL: we prompt Claude-4.5-Sonnet~\citep{anthropic2025sonnet45} to generate multi-turn reasoning trajectories for the 60 seed questions used in our synthetic generation pipeline, then finetune the base model on these interactions via standard SFT.
And \texttt{\ours} denotes models finetuned via trajectory-wise GRPO on our synthetic corpus, starting from the \texttt{Base} checkpoint.
\texttt{SFT-\ours} applies the same GRPO training but initializes from the Seed-SFT checkpoint, allowing us to assess whether SFT and RL provide complementary benefits.
Additionally, we report the performance of Claude-4.5-Sonnet, DeepSeek-V3.1~\citep{liu2024deepseek}, and Gemini-2.5-Flash~\citep{huang2025gemini} as reference points from models that are orders of magnitude larger.
%

\begin{wrapfigure}{r}{0.65\textwidth}
\vspace{-0.15in}
    \centering
    \includegraphics[width=\linewidth]{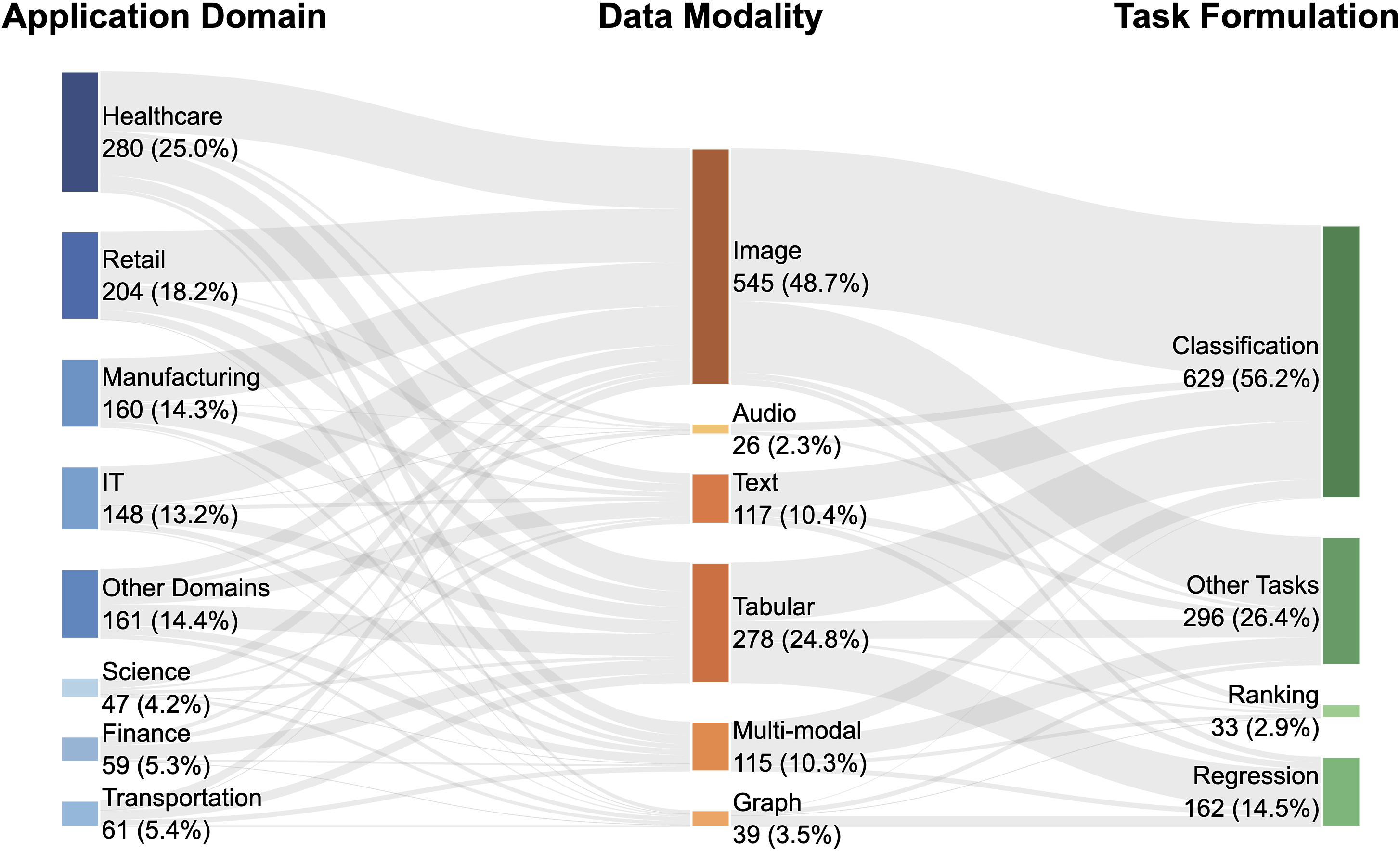}
    \vspace{-0.2in}
    \caption{
        Distribution of the synthetic training data across three axes: application domain (left), data modality (center), and task formulation (right). 
        %
    }
    \label{fig:data_analysis}
\vspace{-0.2in}
\end{wrapfigure}

\paragraph{Implementation Details.}
Given the milestone-to-tier mapping described above, we assign the specific reward weights in our dense reward formulation \secref{subsec:trajectory_grpo} as follows: $w_{\text{format}} = 0.1$, $w_{\text{execute}} = 0.3$, $w_{s_\text{median}} = 0.1$, $w_{s_\text{bronze}} = 0.2$, $w_{s_\text{silver}} = 0.2$, and $w_{s_\text{gold}} = 0.1$, weighting the higher-tier milestones more heavily to incentivize competitive performance. 
To accommodate the context window capacity of Qwen3, we restrict the maximum interaction limit to $T_{\max} = 20$ and the execution time limit of $\tau_{\max} = 90$ seconds in GRPO training.
Additional hyperparameter details can be found in \appref{sec:implementation}.

\subsection{Statistics of Synthetic Training Data}
\label{sec:statistics}
\paragraph{Domain, Modality, and Tasks.}
A key design goal of our synthetic generation pipeline is to produce a training curriculum that exposes the RL agent to a wide spectrum of machine learning scenarios.
As shown in \figref{fig:data_analysis}, the resulting corpus exhibits substantial diversity along three axes.
In terms of application domain, the tasks span healthcare (25.0\%), retail (18.2\%), manufacturing (14.3\%), IT (13.2\%), and several additional sectors including transportation, finance, and science. 
For data modality, image-based tasks (48.7\%) and tabular tasks (24.8\%) constitute the majority, complemented by text (10.4\%), multi-modal (10.3\%), graph (3.5\%), and audio (2.3\%) tasks, ensuring the agent encounters a range of feature representations and preprocessing requirements. 
Finally, the task formulation distribution is anchored by classification (56.2\%) but also covers regression (14.5\%), ranking (2.9\%), and other formulations such as forecasting and reconstruction (26.4\%).
This diversity is notable given that the entire corpus is procedurally derived from only 60 seed tasks, demonstrating the amplification capacity of the \texttt{Data Strategist} agent described in \secref{subsec:synthetic_pipeline}.

\begin{wrapfigure}{r}{0.45\textwidth}
    \centering
    \includegraphics[width=\linewidth]{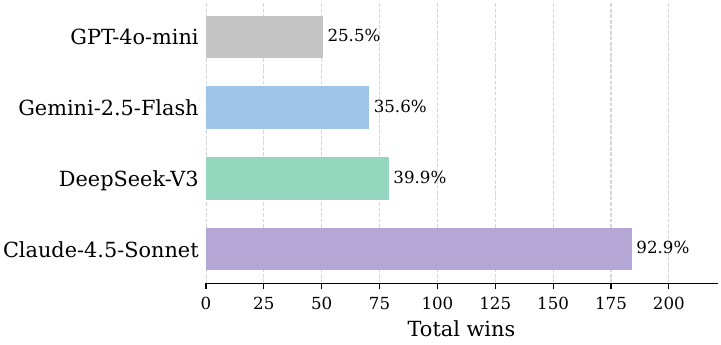}
    \vspace{-0.25in}
    \caption{
        Pairwise win counts and win rates (\%) across 64 synthetic tasks. 
        Each model is evaluated against the other three; ties count as 0.5 wins. 
    }
    \label{fig:model_pairwise}
    \vspace{-0.25in}
\end{wrapfigure}

\paragraph{Task Reliability and Difficulty Calibration.}
To validate that our synthetic tasks are both non-trivial and properly calibrated in difficulty, we benchmark four models with well-established capability differences: \texttt{GPT-4o-mini}~\citep{hurst2024gpt}, \texttt{Gemini-2.5-Flash}~\citep{huang2025gemini}, \texttt{DeepSeek-V3}~\citep{guo2025deepseek}, and \texttt{Claude-4.5-Sonnet}~\citep{anthropic2025sonnet45}.

\begin{wrapfigure}{r}{0.45\textwidth}
    \centering
    \includegraphics[width=\linewidth]{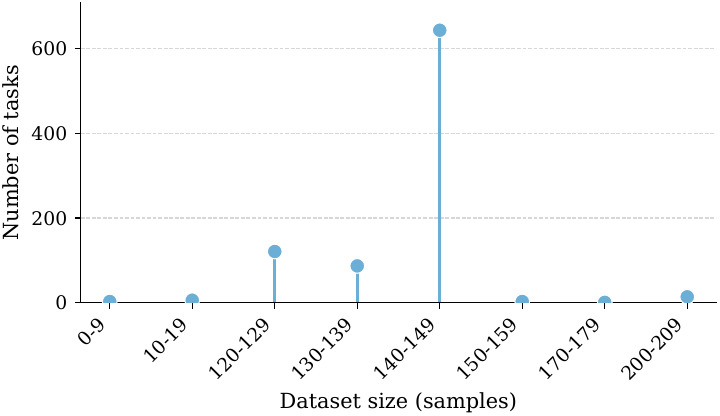}
    \vspace{-0.25in}
    \caption{
        Distribution of per-task dataset sizes across the synthetic training corpus.
    }
    \label{fig:data_size_distribution}
    \vspace{-0.1in}
\end{wrapfigure}

We randomly sample 64 tasks from our synthetic corpus and compare model performances pairwise, assigning 1 point for a win, 0 for a loss, and 0.5 for a tie.
As shown in \figref{fig:model_pairwise}, the aggregated win counts reproduce the expected capability ordering: Claude-4.5-Sonnet dominates with a 92.9\% win rate, followed by DeepSeek-V3 (39.9\%), Gemini-2.5-Flash (35.6\%), and GPT-4o-mini (25.5\%).
The fact that our synthetic tasks cleanly separate models of known capability confirms that they capture meaningful MLE task difficulty and are sufficiently challenging to serve as a reliable training signal.

\paragraph{Dataset Scale and Execution Latency.}
Each synthetic task generated by our pipeline comes with an associated training dataset that the agent's ML pipeline must process during rollouts.

\begin{wrapfigure}{r}{0.45\textwidth}
    \centering
    \includegraphics[width=\linewidth]{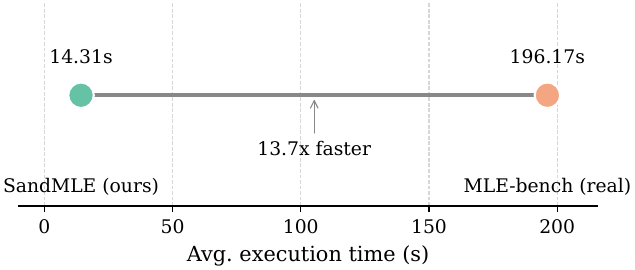}
    \vspace{-0.25in}
    \caption{
        Avg. code execution time on original MLE-bench seed tasks versus \ours synthetic tasks.
    }
    \label{fig:time_comp}
\end{wrapfigure}
The size of this per-task dataset directly determines code execution latency and is therefore the key lever for making on-policy RL feasible.
For context, the original 60 seed tasks from MLE-bench contain an average of approximately 4.09 million samples per task~\citep{chan2024mle}, making each rollout step prohibitively slow. 
By contrast, our pipeline intentionally constrains the per-task dataset to a micro-scale regime: as shown in \figref{fig:data_size_distribution}, the majority of tasks contain between 120 and 150 samples each.

The impact on execution latency is dramatic.
Using code implementations generated by \texttt{Gemini-2.5-Flash}~\citep{huang2025gemini}, the average execution time drops from 196.17 seconds on the original MLE-bench tasks to just 14.31 seconds on our synthetic tasks, a reduction of over $13\times$, as shown in \figref{fig:time_comp}.
This speedup is what transforms trajectory-wise GRPO from infeasible to practical, enabling thousands of on-policy rollout updates in the wall-clock time previously consumed by a handful of steps on the original benchmark.
A comprehensive qualitative analysis of a representative generated task, illustrating the complexity and mathematical rigor preserved despite this scale reduction, is provided in \appref{sec:qualitative_analysis}.

\subsection{Main Results}\label{sec:main_results}

\paragraph{Comparison with Baselines.}
As shown in \tabref{tab:main_results}, \texttt{\ours} achieves significant gains at every scale: +66.9\% relative improvement on \texttt{Any Medal} rate at 8B, +24.7\% at 14B, and +100.7\% at 30B over their respective \texttt{Base} models. 
By comparison, the off-the-shelf \texttt{Base} models achieve between 13.6\% and 18.2\% \texttt{Any Medal} rate, and the \texttt{Seed-SFT} baseline---despite leveraging 60 high-quality \texttt{Claude-4.5-Sonnet} trajectories---yields marginal or zero improvement over \texttt{Base} (e.g., \texttt{Any Medal} rate remains at 13.6\% for both 8B and 30B). In contrast, our \texttt{\ours} achieves a substantial 20.3\% to 66.9\% relative improvement over the \texttt{Seed-SFT} baseline.
This gap suggests that behavioral cloning on expert trajectories does not transfer the iterative problem-solving behavior that trajectory-wise RL acquires through direct environment interaction.
Moreover, these RL-driven improvements propel our open-weight models to rival advanced closed-source systems; our 8B \texttt{\ours} matches the 22.7\% \texttt{Any Medal} rate of Deepseek-V3.1 and Gemini-2.5-flash, while our 14B and 30B models (27.3\%) aggressively close the gap with the top-performing Claude-4.5-Sonnet (31.8\%).

\paragraph{Combining SFT and RL.}
Initializing GRPO from the \texttt{Seed-SFT} checkpoint (\texttt{SFT-\ours}) consistently improves operational reliability without sacrificing \texttt{Any Medal} rate. 
as shown in \tabref{tab:main_results}, at 8B, \texttt{SFT-\ours} raises \texttt{Valid Submission} from 63.6\% to 90.9\% while maintaining 22.7\% \texttt{Any Medal} rate.
At 14B, this variant achieves the strongest overall profile at its scale: 95.5\% \texttt{Valid Submission} and 27.3\% \texttt{Any Medal}. 
These results indicate that SFT and RL contribute along different axes---format compliance and pipeline construction from SFT, higher-order reasoning from RL---and combine effectively.

\paragraph{Scaling with Model Sizes.}
Scaling model capacity yields improvements along multiple axes. 
Most visibly, \texttt{Valid Submission} for pure \texttt{\ours} increases monotonically from 63.6\% (8B) to 77.3\% (14B) to 100\% (30B), while the \texttt{Base} models remain flat at 68--73\% across scales. 
This suggests that, at smaller scales, the RL policy explores aggressive strategies that do not always produce valid outputs, and that sufficient model capacity resolves this tension between exploration and output reliability. 
Beyond submission validity, \texttt{Above Median} rate for \texttt{\ours} also scales from 27.3\% (8B) to 36.4\% (30B), compared to a static 18.2\% for all \texttt{Base} models, indicating that the gains are not limited to a few tasks but reflect broadly improved engineering competence. 
Finally, the reliance on SFT initialization diminishes with scale: the \texttt{Valid Submission} gap between \texttt{\ours} and \texttt{SFT-\ours} narrows from 27.3\% at 8B to 18.2\% at 14B, and inverts at 30B where pure \texttt{\ours} achieves 100\% without SFT. 
This suggests that larger models can internalize both the formatting discipline and the reasoning capability through RL alone.
Additionally, we also conduct an ablation study to our reward design in Appendix \ref{sec:ablation_reward}, which demonstrates the effectiveness of our milestone-based reward in enabling these optimizations.

\begin{table}[t]
\centering
\caption{
    Main results on the MLE-Bench-Lite dataset using the ReAct framework. 
    Performance is reported as the percentage (\%). 
    Best results within each model scale are highlighted in \textbf{bold}.
    \texttt{\ours} refers to model finetuned through our proposed pipeline.
}
\label{tab:main_results}
\setlength{\tabcolsep}{8pt}
\resizebox{\columnwidth}{!}{%
\begin{tabular}{l cccccc}
\toprule
\textbf{Model} & \textbf{Valid Sub.} & \textbf{Above Median} & \textbf{Bronze} & \textbf{Silver} & \textbf{Gold} & \textbf{Any Medal} \\
\midrule
Deepseek-V3.1 & 90.9 & 36.4 & 0.0 & 4.5 & 18.2 & 22.7 \\
Gemini-2.5-flash & 81.8 & 45.5 & 4.5 & 4.5 & 13.6 & 22.7 \\
Claude-4.5-Sonnet & 95.5 & 31.8 & 0.0 & 4.5 & 27.3 & 31.8 \\
\midrule
\multicolumn{7}{c}{\textbf{Qwen3-8B}} \\
\midrule
Qwen3-8B-Base & 68.2 & 18.2 & 0.0 & 4.5 & 9.1 & 13.6 \\
Qwen3-8B-Seed-SFT & 72.7 & 18.2 & \textbf{4.5} & 0.0 & 9.1 & 13.6 \\
Qwen3-8B-\ours & 63.6 & 27.3 & 0.0 & \textbf{4.5} & \textbf{18.2} & \textbf{22.7} \\
Qwen3-8B-SFT-\ours & \textbf{90.9} & \textbf{31.8} & \textbf{4.5} & 0.0 & \textbf{18.2} & \textbf{22.7} \\
\midrule
\multicolumn{7}{c}{\textbf{Qwen3-14B}} \\
\midrule
Qwen3-14B-Base & 72.7 & 18.2 & 4.5 & 4.5 & 18.2 & 18.2 \\
Qwen3-14B-Seed-SFT & 72.7 & 22.7 & 4.5 & 0.0 & 13.6 & 18.2 \\
Qwen3-14B-\ours & 77.3 & 27.3 & \textbf{4.5} & \textbf{4.5} & 13.6 & 22.7 \\
Qwen3-14B-SFT-\ours & \textbf{95.5} & \textbf{31.8} & \textbf{4.5} & 0.0 & \textbf{22.7} & \textbf{27.3} \\
\midrule
\multicolumn{7}{c}{\textbf{Qwen3-30B-A3B}} \\
\midrule
Qwen3-30B-Base & 68.2 & 18.2 & 4.5 & 0.0 & 9.1 & 13.6 \\
Qwen3-30B-Seed-SFT & 77.3 & 31.8 & 4.5 & 4.5 & 13.6 & 22.7 \\
Qwen3-30B-\ours & \textbf{100.0} & 36.4 & 0.0 & \textbf{13.6} & \textbf{13.6} & \textbf{27.3} \\
Qwen3-30B-SFT-\ours & 90.9 & \textbf{45.5} & \textbf{4.5} & 9.1 & \textbf{13.6} & \textbf{27.3} \\
\bottomrule
\end{tabular}
}
\end{table}

\begin{table}[t]
\centering
\caption{Generalization results on the MLE-Bench-Lite dataset across alternative agent frameworks (AIDE, AIRA, ML-Master). Only 14B and 30B models are shown. Best results within each grouping are highlighted in \textbf{bold}.}
\label{tab:generalization_lite}
\setlength{\tabcolsep}{8pt}
\resizebox{\columnwidth}{!}{%
\begin{tabular}{ll cccccc}
\toprule
\textbf{Model} & \textbf{Framework} & \textbf{Valid Sub.} & \textbf{Above Med.} & \textbf{Bronze} & \textbf{Silver} & \textbf{Gold} & \textbf{Any Medal} \\
\midrule
\multicolumn{8}{c}{\textbf{Qwen3-14B}} \\
\midrule
Qwen3-14B-Base & AIDE & 100.0 & 36.4 & 0.0 & 4.5 & \textbf{22.7} & 27.3 \\
Qwen3-14B-Seed-SFT & AIDE & 100.0 & 31.8 & 0.0 & 9.1 & 9.1 & 18.2 \\
Qwen3-14B-\ours & AIDE & \textbf{100.0} & \textbf{40.9} & 0.0 & \textbf{13.6} & 18.2 & \textbf{31.8} \\
\cmidrule{1-8}
Qwen3-14B-Base & AIRA & 100.0 & 22.7 & 0.0 & 0.0 & 9.1 & 9.1 \\
Qwen3-14B-Seed-SFT & AIRA & 95.5 & 22.7 & 0.0 & 0.0 & 13.6 & 13.6 \\
Qwen3-14B-\ours & AIRA & \textbf{100.0} & \textbf{36.4} & \textbf{4.5} & 0.0 & \textbf{18.2} & \textbf{22.7} \\
\midrule
\multicolumn{8}{c}{\textbf{Qwen3-30B-A3B}} \\
\midrule
Qwen3-30B-Base & AIDE & 95.5 & 18.2 & 4.5 & 0.0 & 9.1 & 13.6 \\
Qwen3-30B-Seed-SFT & AIDE & 100.0 & 18.2 & 4.5 & 0.0 & \textbf{9.1} & 13.6 \\
Qwen3-30B-\ours & AIDE & \textbf{100.0} & \textbf{18.2} & \textbf{9.1} & \textbf{0.0} & 4.5 & \textbf{13.6} \\
\cmidrule{1-8}
Qwen3-30B-Base & AIRA & 86.4 & 27.3 & 0.0 & 0.0 & 18.2 & 18.2 \\
Qwen3-30B-Seed-SFT & AIRA & 100.0 & 22.7 & 0.0 & \textbf{4.5} & 9.1 & 13.6 \\
Qwen3-30B-\ours & AIRA & \textbf{100.0} & \textbf{40.9} & \textbf{4.5} & 0.0 & \textbf{22.7} & \textbf{27.3} \\
\bottomrule
\end{tabular}
}
\end{table}

\subsection{Test-Time Scaling of \ours Models}\label{sec:test_time_scaling}

\begin{wrapfigure}{r}{0.5\textwidth}
\vspace{-0.35in}
    \centering
    \includegraphics[width=\linewidth]{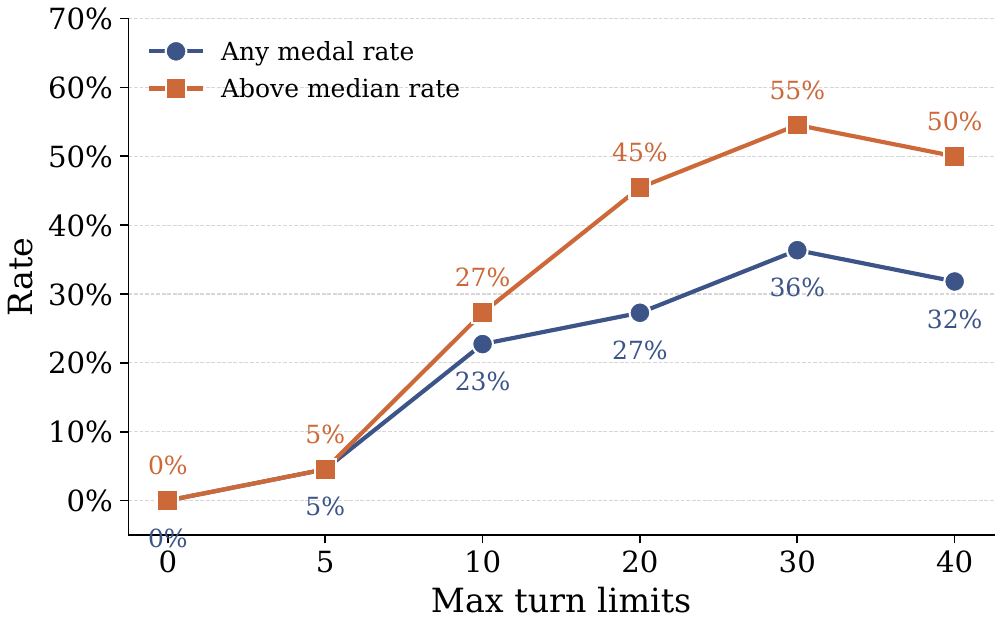}
    \vspace{-0.25in}
    \caption{
        Test-time scaling results of \texttt{Qwen3-30B-SFT-\ours} on MLE-Bench-Lite with ReAct framework.
    }
    \label{fig:test_time_scaling}
    \vspace{-0.15in}
\end{wrapfigure}

To further understand the behavior of our model trained via \ours, we investigate test-time scaling by varying the maximum allowed interaction turns $T_{\max}$ for the \texttt{Qwen3-30B-SFT-\ours} model within the ReAct framework.
For these experiments, the maximum compute limit is fixed at 24 GPU hours per task. 
As the number of iterations $T_{\max}$ grows, the agent's trajectory naturally risks exceeding the context window of the \texttt{Qwen3} model.
To mitigate this, we employ a dynamic truncation strategy: if the context window limit is reached, we systematically remove older messages that contain failed code executions to free up context space. 
As shown in \figref{fig:test_time_scaling}, the model exhibits clear positive test-time scaling, successfully sustaining self-improvement abilities as it is granted more turns to iterate. 
This sustained scaling demonstrates the effectiveness of our \ours approach in instilling robust trial-and-error reasoning. 
Both \texttt{Any Medal} and \texttt{Above Median} rates increase steadily, peaking at a limit of 30 turns. 
However, scaling beyond this threshold causes a performance regression due to frequent context window overflows. 
Despite our message eviction strategy, the sheer loss of historical context disrupts the agent's long-horizon memory, often trapping it in repetitive loops. 
This highlights that test-time scaling in complex MLE tasks is ultimately bottlenecked by the model's effective context length.

\subsection{Training Dynamics}\label{sec:training_dynamics}
We track training and validation rewards across the \texttt{Qwen3-8B, 14B, and 30B} models over GRPO steps, as detailed in \appref{app:training_dynamics}.
We observe that the RL objective converges stably on our synthetic environments across all scales, validating that the synthetic tasks provide a sufficiently diverse and well-calibrated training signal.
Larger models both start higher and converge to higher reward ceilings, consistent with the evaluation-time scaling trends as shown in \tabref{tab:main_results}.
Notably, the ability to produce valid submissions is itself a learned behavior: smaller models acquire it slowly and inconsistently during training, while the bigger 30B model internalizes it early and maintains it throughout, explaining why pure \texttt{Qwen3-30B-\ours} no longer requires SFT initialization for output reliability.

\section{Analysis}\label{sec:generalization}
\subsection{Framework Generalization}
%
Since model performance on MLE tasks is sensitive to the choice of agent scaffold, we evaluate whether the gains from \ours transfer beyond the ReAct scaffold used during training. 
Specifically, we test the 14B and 30B models on MLE-Bench-Lite~\citep{chan2024mle} with AIDE~\citep{jiang2025aide} and AIRA~\citep{toledo2025ai}, and on MLE-Dojo with AIDE~\citep{jiang2025aide} and MLE-Agent~\citep{qiang2025mle}.
%

\begin{table}[t]
\centering
\caption{
    Generalization results on the MLE-Dojo benchmark. 
    The HumanRank Score measures relative leaderboard performance against human competitors. 
}
\label{tab:generalization_dojo}
\setlength{\tabcolsep}{24pt}
\resizebox{0.9\linewidth}{!}{%
\begin{tabular}{ll cc}
\toprule
\textbf{Model} & \textbf{Framework} & \textbf{Valid Sub. (\%)} & \textbf{HumanRank Score} \\
\midrule
Claude-4.5-Sonnet & MLE-Agent & 91.9 & 54.09 \\
\midrule
\multicolumn{4}{c}{\textbf{Qwen3-14B}} \\
\midrule
Qwen3-14B-Base & MLE-Agent & 27.4 & 9.62 \\
Qwen3-14B-Seed-SFT & MLE-Agent & 3.2 & 0.65 \\
Qwen3-14B-\ours & MLE-Agent & \textbf{40.3} & \textbf{12.55} \\
\cmidrule{1-4}
Qwen3-14B-Base & AIDE & 74.2 & 37.73 \\
Qwen3-14B-Seed-SFT & AIDE & 62.9 & 27.51 \\
Qwen3-14B-\ours & AIDE & \textbf{75.8} & \textbf{37.86} \\
\midrule
\multicolumn{4}{c}{\textbf{Qwen3-30B-A3B}} \\
\midrule
Qwen3-30B-Base & MLE-Agent & 71.0 & 29.12 \\
Qwen3-30B-Seed-SFT & MLE-Agent & 17.7 & 7.34 \\
Qwen3-30B-\ours & MLE-Agent & \textbf{83.9} & \textbf{38.56} \\
\cmidrule{1-4}
Qwen3-30B-Base & AIDE & 59.7 & 25.99 \\
Qwen3-30B-Seed-SFT & AIDE & 66.1 & 28.14 \\
Qwen3-30B-\ours & AIDE & \textbf{77.4} & \textbf{28.72} \\
\bottomrule
\end{tabular}
}
\end{table}

As shown in \tabref{tab:generalization_lite} and \tabref{tab:generalization_dojo}, \ours consistently exceeds \texttt{Base} model performance across all scaffold-benchmark combinations.
While the SFT models prove brittle when deployed outside the specific scaffold used during their data generation, most notably collapsing to a 17.7\% \texttt{Valid Submission} rate on MLE-Dojo for the 30B model with MLE-Agent, the \ours models adapt robustly. 
Specifically, on the MLE-Dojo benchmark, Qwen3-30B-\ours paired with MLE-Agent framework achieves an 83.9\% \texttt{Valid Submission} rate and a HumanRank score of 38.56, vastly outperforming both the Base and SFT variants. 
We observe similar robustness on MLE-Bench-Lite, where the 14B and 30B \ours models consistently elevate the \texttt{Any Medal} and \texttt{Valid Submission} rate across AIDE and AIRA scaffolds. 
%

\subsection{Effectiveness of Milestone-Based Rewards}
\label{sec:ablation_reward}
To validate the necessity of our dense, milestone-based reward function (detailed in \secref{subsec:trajectory_grpo}), we conduct an ablation study comparing it against a sparse reward formulation. In complex, long-horizon tasks like machine learning engineering, relying solely on ultimate performance metrics often creates an insurmountable exploration challenge for the RL agent. To demonstrate this, we configure a \texttt{Sparse Reward} baseline where the environmental feedback is restricted strictly to basic formatting compliance and the highest performance standard: $r = 0.1 \cdot r_{\text{format}} + 0.9 \cdot \mathbb{I}_{s_\text{gold}}$. We evaluate this sparse formulation across the 8B, 14B, and 30B model scales using the ReAct framework on MLE-bench-lite.

The results, presented in Table \ref{tab:ablation_reward}, clearly demonstrate the superiority of providing dense, hierarchical feedback. Under the sparse reward regime, the models struggle to find meaningful gradients for improvement, as achieving a $s_\text{gold}$ standard purely through initial exploration is exceedingly rare. Consequently, the sparse models regress significantly in overall performance. 

This degradation is most pronounced at the 30B scale. When deprived of intermediate stepping stones, the Qwen3-30B model's \texttt{Any Medal} rate collapses from 27.3\% (with our milestone reward) to just 13.6\% (with the sparse reward), and its \texttt{Above Median} rate halves from 36.4\% to 18.2\%. Furthermore, the lack of incremental execution rewards (such as successfully generating a valid submission file) destabilizes the agent's baseline coding reliability; the 30B valid submission rate drops from a perfect 100.0\% down to 86.4\%. At the 8B and 14B scales, the sparse reward fails to lift the models beyond the performance of the non-RL SFT baselines (refer back to Table \ref{tab:main_results}), yielding only 13.6\% and 18.2\% medal rates, respectively. 

These findings empirically confirm that a hierarchical reward landscape—which independently validates format, execution, and progressive performance tiers—is essential for stabilizing policy optimization and effectively scaffolding an LLM's transition toward state-of-the-art engineering capabilities.
\begin{table}[t]
\centering
\caption{
    Ablation study on reward design using the MLE-bench-lite dataset.   
    \texttt{Sparse Reward} relies only on format and the highest performance tier ($s_\text{gold}$), whereas our \texttt{\ours} uses the proposed dense, milestone-based reward.
}
\label{tab:ablation_reward}
\setlength{\tabcolsep}{6pt}
\resizebox{\columnwidth}{!}{%
\begin{tabular}{ll cccccc}
\toprule
\textbf{Model} & \textbf{Reward Type} & \textbf{Valid Sub.} & \textbf{Above Median} & \textbf{Bronze} & \textbf{Silver} & \textbf{Gold} & \textbf{Any Medal} \\
\midrule
\multirow{2}{*}{Qwen3-8B}
& Sparse Reward & 59.1 & 22.7 & 0.0 & \textbf{4.5} & 9.1 & 13.6 \\
& Dense Reward (\ours) & \textbf{63.6} & \textbf{27.3} & 0.0 & \textbf{4.5} & \textbf{18.2} & \textbf{22.7} \\
\midrule
\multirow{2}{*}{Qwen3-14B}
& Sparse Reward & \textbf{77.3} & \textbf{31.8} & \textbf{4.5} & 0.0 & \textbf{13.6} & 18.2 \\
& Dense Reward (\ours) & \textbf{77.3} & 27.3 & \textbf{4.5} & \textbf{4.5} & \textbf{13.6} & \textbf{22.7} \\
\midrule
\multirow{2}{*}{Qwen3-30B}
& Sparse Reward & 86.4 & 18.2 & \textbf{4.5} & 4.5 & 4.5 & 13.6 \\
& Dense Reward (\ours) & \textbf{100.0} & \textbf{36.4} & 0.0 & \textbf{13.6} & \textbf{13.6} & \textbf{27.3} \\
\bottomrule
\end{tabular}
}
\end{table}

\section{Conclusion}
\label{sec:conclusion}
In this work, we introduced \ours, a framework that makes trajectory-wise on-policy RL practical for MLE agents by generating diverse, verifiable synthetic environments with micro-scale datasets.
Our approach is grounded in the observation that, unlike SWE tasks where execution time is dominated by compilation and test logic, MLE latency is overwhelmingly driven by the size of the datasets that each ML pipeline must process during training and inference.
By constructing tasks from the ground up at 50--200 samples, we reduce per-step execution time by over 13$\times$ while preserving the structural complexity needed for meaningful policy optimization.
Combined with a dense, milestone-based reward formulation, this enables stable GRPO training that yields 20.3\% to 66.9\% relative improvement in medal rate over the SFT baseline across model scales on MLE-bench-lite, and generalizes robustly across unseen agentic scaffolds.
We believe the principle underlying \ours---that synthetic micro-scale environments can serve as effective proxies for real-world MLE tasks---opens a scalable path toward training MLE agents that improve through direct environment interaction rather than imitation of expert behavior.

\clearpage
\bibliography{colm2026_conference}
\bibliographystyle{colm2026_conference}

\clearpage

\appendix
\section*{Appendix}

\section{Supplementary Implementation Details}
\label{sec:implementation}

\subsection{Metric Definition Details}
\label{sec:metric_definition_details}

To rigorously evaluate the performance of our autonomous agents, we utilize a combination of submission validity, threshold-based medal rates, and relative leaderboard rankings. The specific metrics reported in our main results are defined as follows:

\paragraph{MLE-bench-lite Metrics}
For the MLE-bench-lite benchmark, performance is categorized into several discrete thresholds based on the agent's ability to produce valid, high-quality machine learning pipelines:
\begin{itemize}
    \item \textbf{Valid Submission Rate:} The percentage of tasks where the agent successfully completes the task and generates a properly formatted \texttt{submission.csv} file that can be parsed and scored by the evaluation environment without runtime or schema errors.
    \item \textbf{Above Median:} The percentage of tasks where the agent's valid submission achieves a score strictly better than the median (50th percentile) performance of the original human leaderboard for that specific competition.
    \item \textbf{Bronze}, \textbf{Silver}, and \textbf{Gold:} The percentage of tasks where the agent's submission achieves a score that meets or exceeds the respective Kaggle medal thresholds (Bronze, Silver, or Gold) established for that competition.
    \item \textbf{Any Medal:} The overall percentage of tasks where the agent achieves at least a Bronze-level performance (i.e., the union of Bronze, Silver, and Gold successes). This serves as our primary metric for high-level competency.
\end{itemize}

\paragraph{MLE-Dojo Metrics}
For the MLE-Dojo benchmark, we evaluate the agent's continuous relative performance against human participants using the \textbf{HumanRank Score}. The HumanRank score measures the relative ranking of an agent's submission within the historical competition leaderboard. Submissions receive higher scores if they achieve a better rank among all participants. 

Suppose that an agent's submission ranks at position $p$ among a total of $N$ submissions on the leaderboard. The position score $s$ is computed as:
$$ s = 1 - \frac{p}{N} $$
To prevent evaluation bias and account for distribution shifts between public and private test sets, we compute the relative position score $s$ on the public and private leaderboards independently. The final HumanRank score assigned to the agent is the average of these two independent scores.

\newpage
\subsection{Hyperparameter Setup}
We implement our GRPO training pipeline using the RLLM framework \citep{rllm2025}. During the actor rollout phase, we sample candidate trajectories with a group size of $n=4$ at a generation temperature of 1.0. The models are trained for 100 steps with a learning rate of $1 \times 10^{-6}$ and a batch size of 16. For the GRPO objective, we apply a clipping ratio of 0.28 and disable KL divergence penalties. To manage computational resources, each task is allocated a single NVIDIA H200 GPU. We enforce a strict single-step execution time limit of $\tau_{\max} = 90$ seconds during GRPO rollouts, which is extended to $\tau_{\max} = 4$ hours during the final MLE-bench evaluation. During the evaluation phase, we fix the generation temperature to 0.0 to ensure deterministic outputs. 

For our baseline comparisons, we strictly adhere to the configurations and prompts established in their original works~\citep{jiang2025aide, toledo2025ai, qiang2025mle}. For both AIDE and AIRA, experiments are conducted in a single-agent setup powered by one NVIDIA H200 GPU. Each job is constrained by a hard wall-clock cap of 24 hours, an execution time limit of 4 hours, and a 5-minute grace period. The language model generation parameters. For evaluations using MLE-agent scaffolds on the MLE-Dojo benchmark, the total number of interaction steps is limited to 15, with agents maintaining full access to their interaction histories. These evaluation sessions are capped at a maximum runtime of 12 hours on a single NVIDIA H200 GPU. Additionally, the context window is tightly managed, with a maximum input token length of 50,000 and each output round capped at 8,192 tokens.

\newpage
\section{Supplementary Results}
\label{sec:supplementary_results}

\subsection{Training Dynamics}\label{app:training_dynamics}
\begin{figure}[ht]
    \centering
    \begin{subfigure}[b]{0.32\textwidth}
        \centering
        \includegraphics[width=\textwidth]{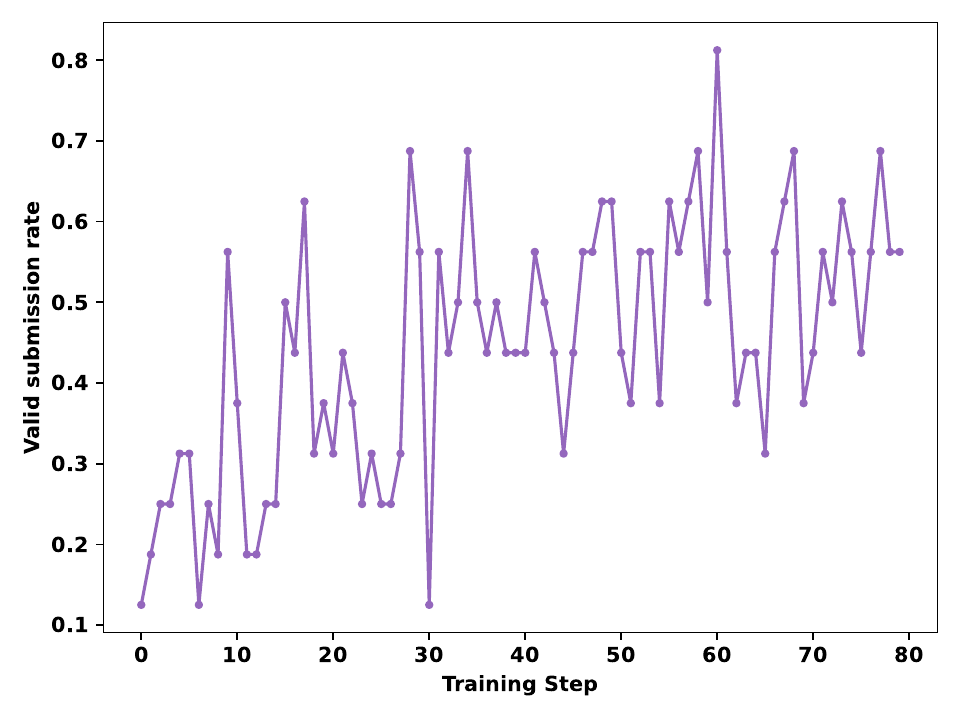}
        \caption{8B Valid Submission Rate}
        \label{fig:8b_submit}
    \end{subfigure}
    \hfill
    \begin{subfigure}[b]{0.32\textwidth}
        \centering
        \includegraphics[width=\textwidth]{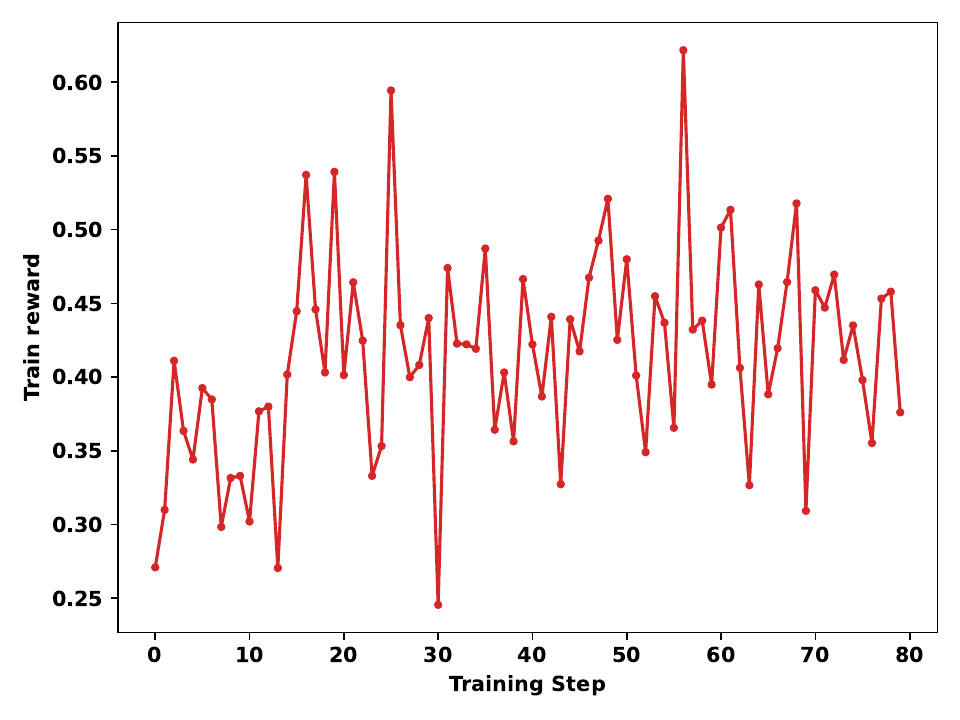}
        \caption{8B Train Reward}
        \label{fig:8b_train_reward}
    \end{subfigure}
    \hfill
    \begin{subfigure}[b]{0.32\textwidth}
        \centering
        \includegraphics[width=\textwidth]{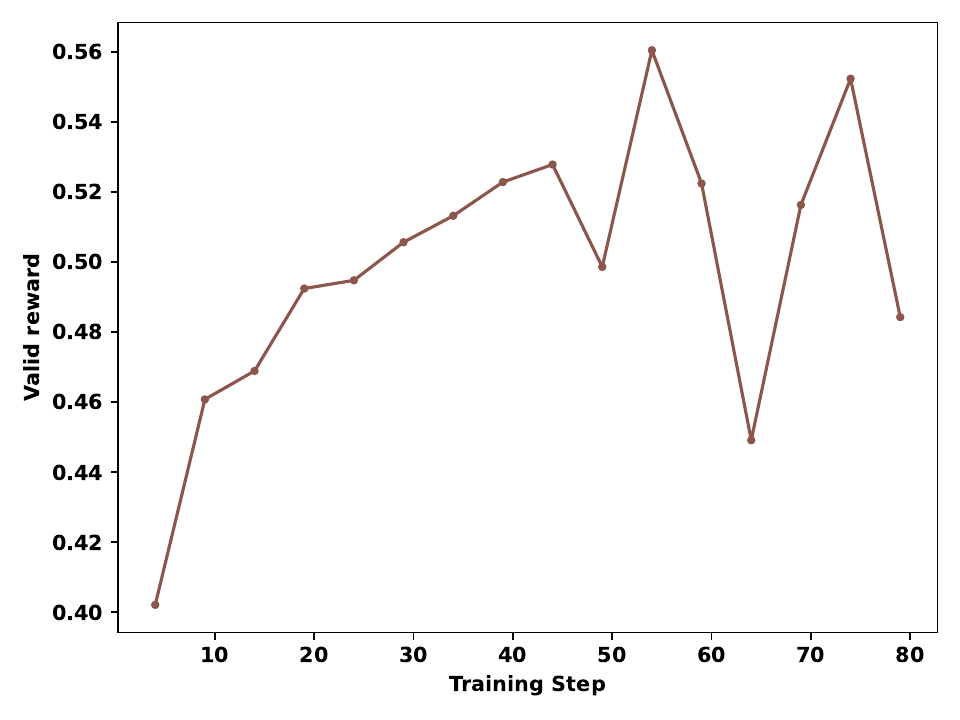}
        \caption{8B Validation Reward}
        \label{fig:8b_val_reward}
    \end{subfigure}
    
    \vspace{0.3cm} 
    
    \begin{subfigure}[b]{0.32\textwidth}
        \centering
        \includegraphics[width=\textwidth]{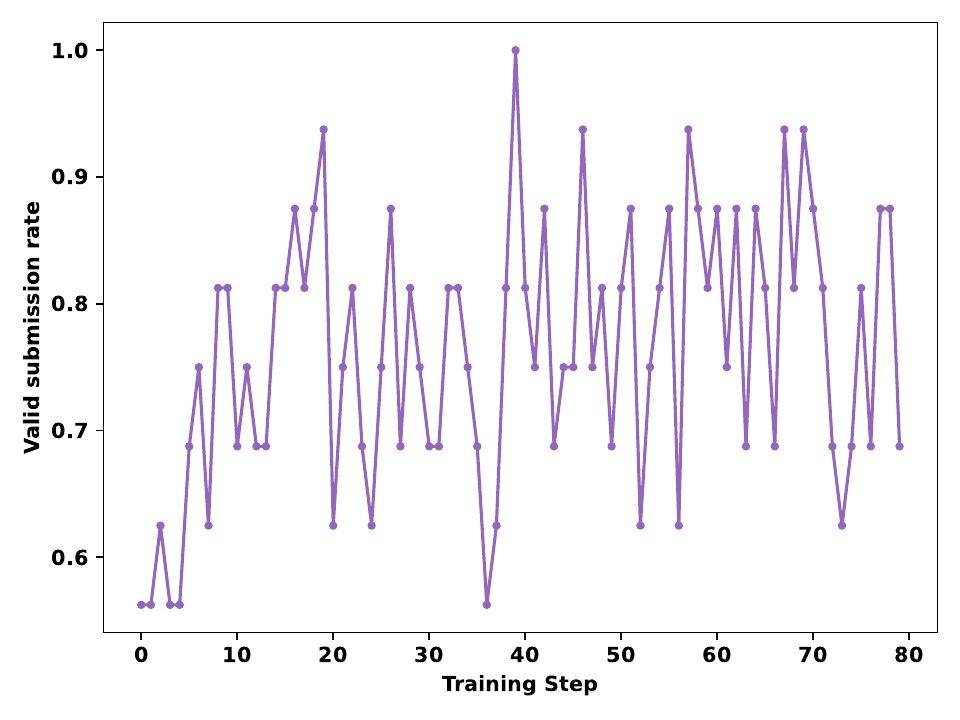}
        \caption{14B Valid Submission Rate}
        \label{fig:14b_submit}
    \end{subfigure}
    \hfill
    \begin{subfigure}[b]{0.32\textwidth}
        \centering
        \includegraphics[width=\textwidth]{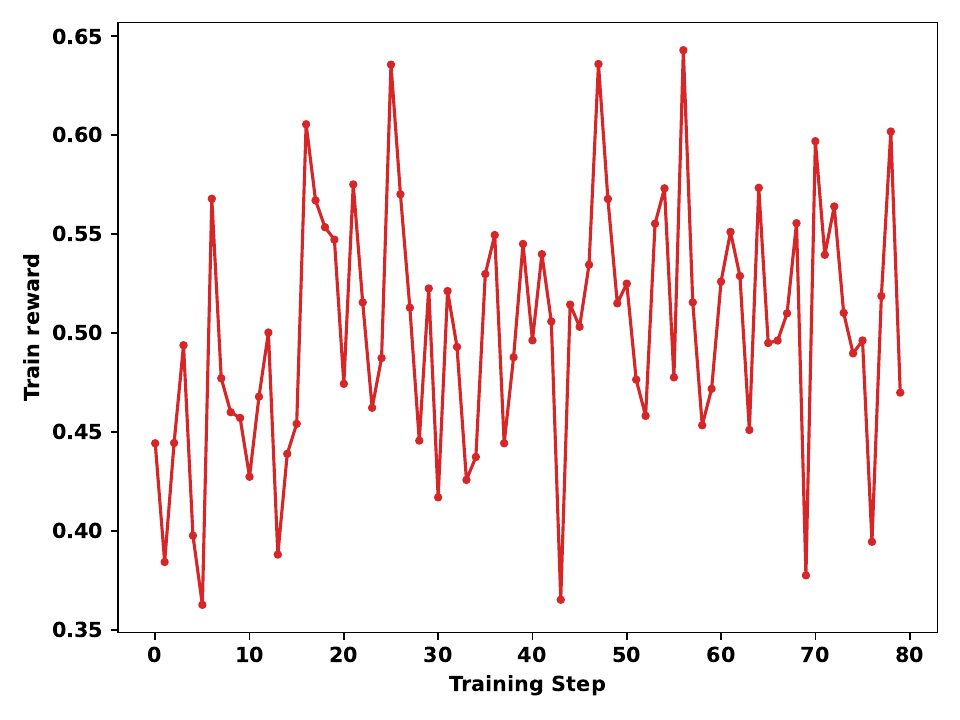}
        \caption{14B Train Reward}
        \label{fig:14b_train_reward}
    \end{subfigure}
    \hfill
    \begin{subfigure}[b]{0.32\textwidth}
        \centering
        \includegraphics[width=\textwidth]{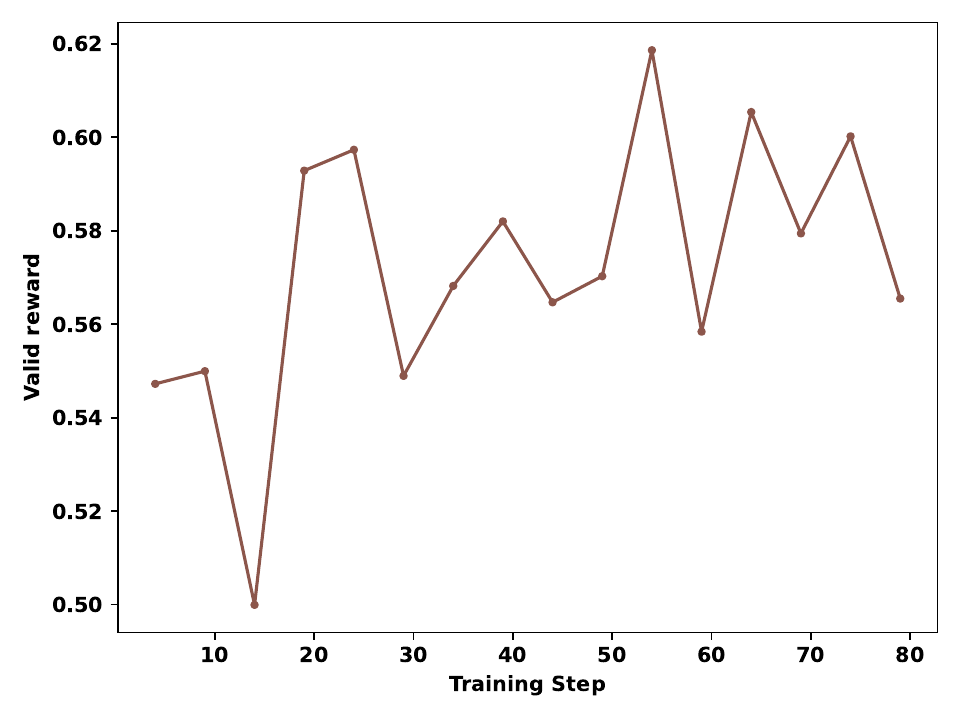}
        \caption{14B Validation Reward}
        \label{fig:14b_val_reward}
    \end{subfigure}

    \begin{subfigure}[b]{0.32\textwidth}
        \centering
        \includegraphics[width=\textwidth]{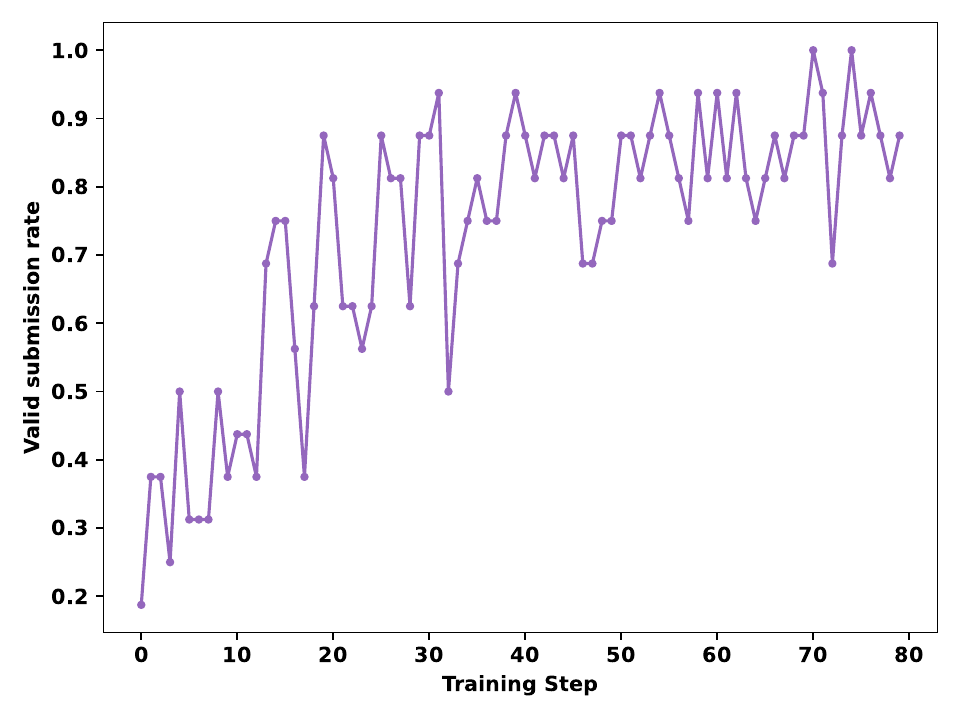}
        \caption{30B Valid Submission Rate}
        \label{fig:30b_submit}
    \end{subfigure}
    \hfill
    \begin{subfigure}[b]{0.32\textwidth}
        \centering
        \includegraphics[width=\textwidth]{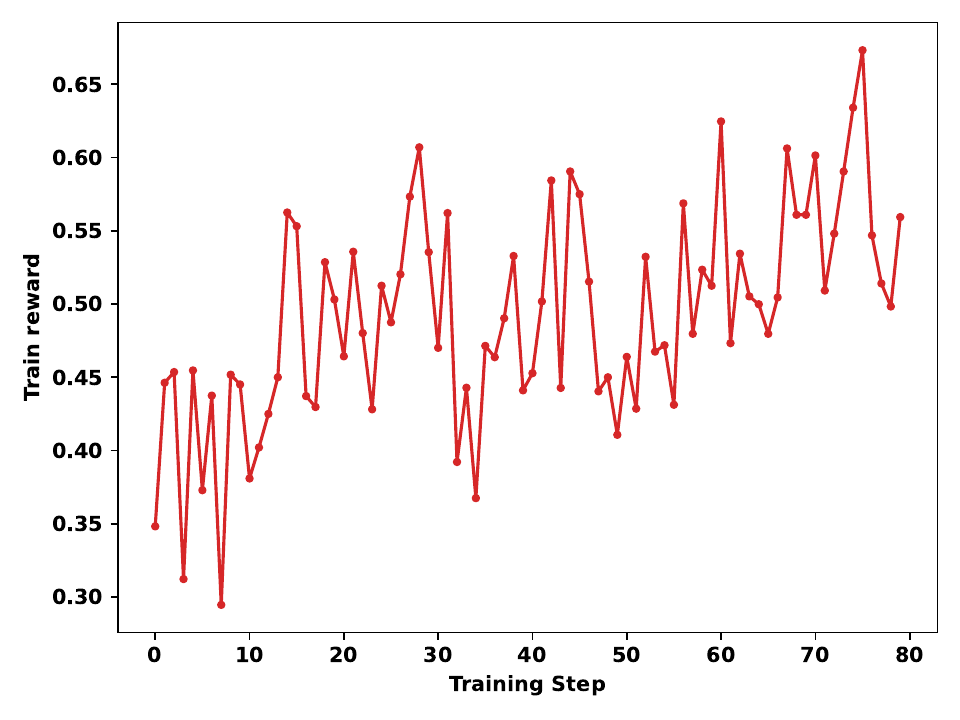}
        \caption{30B Train Reward}
        \label{fig:30b_train_reward}
    \end{subfigure}
    \hfill
    \begin{subfigure}[b]{0.32\textwidth}
        \centering
        \includegraphics[width=\textwidth]{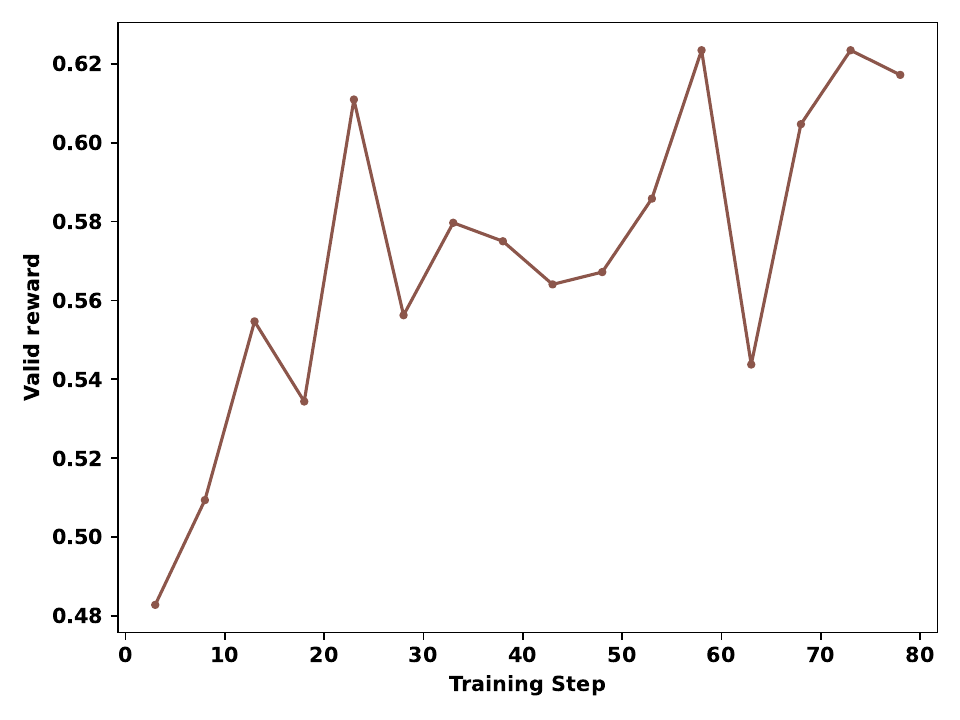}
        \caption{30B Validation Reward}
        \label{fig:30b_val_reward}
    \end{subfigure}
    
    \caption{Training dynamics for the Qwen3-8B (top row), Qwen3-14B (middle row), and Qwen3-30B (bottom row) models optimized via GRPO over 80 steps. The plots display the valid submission rates, training rewards, and validation rewards. All models demonstrate positive learning trends and strong overall convergence, with increased model scale directly correlating to a higher performance ceiling and greater stability, particularly evident in the 30B model's valid submission rate.}
    \label{fig:training_dynamics}
\end{figure}

\figref{fig:training_dynamics} presents the full training curves for the Qwen3-8B (top row), Qwen3-14B (middle row), and Qwen3-30B (bottom row) models optimized via GRPO over 80 steps. 
Each row reports three metrics: valid submission rate, training reward, and validation reward.

All three models demonstrate clear upward trends in reward signals throughout training. 
The validation rewards confirm continuous policy improvement without signs of severe overfitting to the synthetic training environments, as all models stabilize in the final training stages rather than diverging.

The scaling effect of model capacity is most apparent in two aspects. 
First, training rewards: the 30B model achieves higher and more stable training rewards (peaking near 0.67) compared to the 14B model (peaking near 0.62), reflecting stronger initial states and higher performance ceilings with increased capacity. 
Second, valid submission rates: the 8B model fluctuates substantially between 0.1 and 0.8 throughout training; the 14B model reaches 1.0 only intermittently; and the 30B model rapidly climbs to near-perfect submission rates and sustains them consistently in the later stages. 
This progression indicates that larger models more reliably learn to produce well-formatted outputs during RL training, corroborating the evaluation-time results in \tabref{tab:main_results}.
\newpage

\newpage
\section{Supplementary Analysis of the Generated Synthetic Dataset}
\label{sec:qualitative_analysis}

\subsection{Reliability-driven Filtering during Generation}
To ensure the reliability of the synthetic training corpus, the initial pool of tasks undergoes a rigorous multi-stage generation and verification pipeline. 
Following the initial domain and noise amplification phase, we begin with a foundational set of 1200 tasks. 
During the generation of the training data, we allow a maximum of five attempts per task. 
After applying strict execution based verification, this set is refined to 1119 successful tasks. 
Next, we construct the evaluation environments for these remaining tasks, again permitting up to three generation attempts combined with execution based verification, which yields 1106 viable tasks. 
Finally, we perform a comprehensive sanity check on the evaluation results derived from the synthetic evaluation environment, as detailed in \secref{sec:sanity}. 
This concluding filtration step results in a final, high quality corpus of 912 valid tasks, each fully equipped with a structured step-wise feedback environment.

\subsection{Qualitative Analysis}
To demonstrate the efficacy of our multi-agent generation pipeline, we provide a qualitative analysis of a representative synthetic task generated by \ours: \textit{Urban Road Surface Damage Classification Under Motion Blur}.

\textbf{Task Derivation and Mutation Strategy.}
The original seed task (iwildcam-2019-fgvc6 in MLE-bench) focused on classifying images of animal species. The Data Strategist agent extracted the structural DNA of the seed data (a 23-class image classification problem with an extremely long-tail distribution) and applied a deliberate ``noise'' mutation: motion blur. The pipeline explicitly contextualized this mutation by framing the scenario within the smart city setting around vehicle-mounted cameras for road damage detection, where motion blur is a natural, systematic degradation. This mutation preserves the semantic goal and taxonomy of the seed task while significantly shifting the difficulty toward blur-invariant feature extraction and temporal reasoning.

\textbf{Data Generation and Hidden Rules.}
As executed by the MLE Developer agent, the generated dataset strictly adheres to the micro-scale constraints necessary for rapid RL rollouts, comprising only 147 total synthetic RGB images ($1024 \times 768$ resolution) divided into training and testing sets. The generative logic constructs road textures using Perlin noise and overlays specific damage types (e.g., alligator cracking, potholes). Crucially, the mathematical hidden rules explicitly enforce the mutation:
\begin{itemize}
    \item \textbf{Temporal Blur Accumulation:} Images are generated in temporal sequences (1–5 frames). Motion blur is applied via a directional line-kernel, with the blur severity linearly increasing with the frame number to simulate a moving vehicle.
    \item \textbf{Logic-Driven Labels:} The ground-truth labeling is deterministic, driven by edge ratio and contrast for severity tiers, and frequency/texture interactions for specific subclasses. The motion blur acts as structured label noise, deliberately degrading minor damage into the ``intact'' class.
\end{itemize}

\textbf{Evaluation and Alignment.}
The MLOps Engineer constructed a deterministic evaluation sandbox tailored to the specific challenges of the generated data. Recognizing the extreme class imbalance (the ``intact'' class dominates), the environment utilizes Macro-F1 as the evaluation metric to reward performance on rare damage classes. Furthermore, the medal thresholds (e.g., Gold = 0.68, Silver = 0.55) correctly account for the performance degradation caused by the injected blur, requiring an agent to successfully implement deblurring, frequency-domain feature extraction, or sequence aggregation to achieve a winning score. 

Ultimately, this qualitative example illustrates how \ours successfully generates a coherent, domain-consistent environment with complex, verifiable mathematical rules, providing a rigorous testbed for trajectory-wise RL.

\newpage
\section{Prompt Details}

\subsection{React framework}
\label{sec:react_prompt}

\begin{longtable}{p{0.95\textwidth}}
\caption{System and User Prompt templates for the ReAct agent framework.}
\label{tab:react_prompts} \\
\toprule
\textbf{ReAct Framework Prompts} \\
\midrule
\endfirsthead

\multicolumn{1}{c}%
{{\bfseries Table \ref{tab:react_prompts} continued from previous page}} \\
\toprule
\textbf{ReAct Framework Prompts} \\
\midrule
\endhead

\midrule
\multicolumn{1}{r}{{Continued on next page}} \\
\endfoot

\bottomrule
\endlastfoot

\textbf{System Prompt} \newline
\newline
You are an expert Kaggle competitor. Produce one Python script that trains a model and writes \texttt{submission.csv} for the dataset in the user prompt.\newline
\newline
\textbf{Rules:}\newline
- Use only already-installed common libraries (no installs).\newline
- Use the PythonInterpreter tool to iteratively write/run/update your script.\newline
- After producing a submission, use the Score tool to grade it; if the score is unsatisfying, keep refining the code and re-grading until you are satisfied.\newline
- Be concise and task-focused.\newline
\newline
\textbf{Loop:}\newline
1) You are a multi-turn generation agent: in each turn, propose/refine the script or reasoning, then wait for environment/tool feedback.\newline
2) Execute via the tool until it runs cleanly and produces the file. STRICT: each response may contain exactly ONE \texttt{$<$tool\_call$>$} block—do not emit multiple tool calls.\newline
3) After generating the code, the Python environment will provide feedback. You must observe at least one tool feedback (execution result wrapped in \texttt{$<$tool\_response$>$...$<$/tool\_response$>$} tags) before deciding to end. Only when feedback looks good do you reply with \texttt{$<$answer$>$submission$<$/answer$>$}; otherwise continue iterating (do not output \texttt{$<$answer$>$} tags).\newline
4) Use PythonInterpreter to run updated code; use Score tool to grade \texttt{submission.csv}. Repeat this refine-grade loop until the submission is acceptable, then end with \texttt{$<$answer$>$submission$<$/answer$>$}.\newline
\newline
\textbf{Tool usage:}\newline
For each function call, return a json object with function name and arguments within \texttt{$<$tool\_call$>$...$<$/tool\_call$>$} XML tags:\newline
- Wrap executable code exactly like this:\newline
\texttt{$<$tool\_call$>$}\newline
\texttt{python}\newline
\texttt{$<$code$>$}\newline
\texttt{\# Your Python code here}\newline
\texttt{print("Hello World")}\newline
\texttt{$<$/code$>$}\newline
\texttt{$<$/tool\_call$>$}\newline
Code inside those tags runs in Python; keep the tool name \texttt{python} and include \texttt{$<$code$>$...$<$/code$>$}.\newline
\newline
- To grade the \texttt{submission.csv} file, you need to use Score tool and output json object like this:\newline
\texttt{$<$tool\_call$>$}\newline
\texttt{\{"name": "Score", "arguments": \{"competition\_id": "competition id here, such as aerial-cactus-identification, which can be found in the task description"\}\}}\newline
\texttt{$<$/tool\_call$>$}\newline
\newline
Current date: \\

\midrule

\textbf{User Prompt} \newline
\newline
You are solving the task below. Follow the requirements precisely.\newline
\newline
\texttt{\{mle task description\}}\newline
\newline
Your code should adhere to the following requirements:\newline
- Prefer and explicitly use GPU (CUDA) acceleration when available (one A100 GPU should be available): move models/tensors to GPU and handle CPU fallback if CUDA is not present.\newline
- Each Python interpreter execution must finish within a given time limit.\newline
- Overall runtime limits: the agent may take up to {max LLM calls} turns.\newline
- Load train/test data from the provided dataset folder (\texttt{\#\# Dataset Folder}). Please first check the data files and their formats (file types, column names, row counts, etc.).\newline
- Match the exact columns/headers in \texttt{sample\_submission.csv} (\texttt{\#\# Dataset Folder}) and write \texttt{submission.csv} to the \textbf{current directory}.\newline
- Use only common preinstalled libraries (no installs).\newline
- DO NOT display progress bars due to the context window limit. If you have to use function integrated with progress bars, disable progress bars or use the appropriate parameter to silence them.\newline
- Please restrict the use of external libraries to the common libraries.\newline
- The task is an out-of-date competition, so please ignore the timeline in the task description. \\
\end{longtable}

\newpage
\subsection{\ours Method}
\label{sec:prompt}
We provide detailed prompts for \ours in this section.

\begin{longtable}{p{0.95\textwidth}}
\caption{Prompt templates for the Data Strategist agent.}
\label{tab:data_strategist_prompts} \\
\toprule
\textbf{Data Strategist Prompt Template} \\
\midrule
\endfirsthead

\multicolumn{1}{c}%
{{\bfseries Table \ref{tab:data_strategist_prompts} continued from previous page}} \\
\toprule
\endhead

\midrule
\multicolumn{1}{r}{{Continued on next page}} \\
\endfoot

\bottomrule
\endlastfoot

You are an expert Meta-Learning Analyst. Your goal is to extract the ``Structural DNA'' of a machine learning dataset in this folder.\newline
\newline
You must \textbf{IGNORE} the domain context (the ``Story''). For example:\newline
- If the data is about ``Titanic Survivors'', do NOT mention ships, passengers, or icebergs.\newline
- If the data is about ``House Prices'', do NOT mention kitchens, square footage, or neighborhoods.\newline
\newline
Instead, convert these into \textbf{Abstract Mathematical Concepts}.\newline
- ``Age'' $\rightarrow$ ``Continuous variable, positive, right-skewed''.\newline
- ``Cabin Number'' $\rightarrow$ ``High-cardinality categorical, high missing rate''.\newline
- ``Survived'' $\rightarrow$ ``Binary Target, Class Imbalance 60/40''.\newline
\newline
\textbf{CRITICAL INSTRUCTION:}\newline
1. \textbf{Detect the Modality:} Determine if the problem is Tabular, Computer Vision (Images), NLP (Text), Audio or Graph.\newline
2. \textbf{IGNORE Domain:} Do not mention ``X-Rays'', ``Customer Reviews'', or ``Bird Calls''. Describe the signal mathematically.\newline
3. \textbf{Select the Schema:} structure your JSON output according to the detected modality.\newline
\newline
\textbf{OUTPUT SCHEMA (Polymorphic):}\newline
Return ONLY a raw JSON object with this structure:\newline
\texttt{\{}\newline
\hspace*{1em}\texttt{"modality": "Tabular | Image | Text | Audio | Graph",}\newline
\hspace*{1em}\texttt{"task\_type": "Classification | Regression | Segmentation | Object Detection",}\newline
\hspace*{1em}\texttt{"dataset\_stats": \{ "sample\_count": Integer, "is\_imbalanced": Boolean \},}\newline
\hspace*{1em}\texttt{// --- UNIVERSAL TARGET INFO ---}\newline
\hspace*{1em}\texttt{"target\_info": \{}\newline
\hspace*{2em}\texttt{"type": "Label | BoundingBox | Mask | Text",}\newline
\hspace*{2em}\texttt{"cardinality": Integer,}\newline
\hspace*{2em}\texttt{"distribution": "Balanced | Long-tail"}\newline
\hspace*{1em}\texttt{\}}\newline
\texttt{\}} \\
\midrule

You are a Synthetic Data Strategist. I will provide the ``Structural DNA'' of a dataset.\newline
Your goal is to brainstorm 5 distinct ``Industry Scenarios'' that could naturally generate data with this exact structure.\newline
\newline
\textbf{INPUT DNA:}\newline
\texttt{\{dna\_json\}}\newline
\newline
\textbf{CONSTRAINT:}\newline
The scenarios must strictly justify the features:\newline
- If DNA has ``Long Text'', the domain must involve documents/logs/dialogue.\newline
- If DNA has ``Images'', the domain must involve visual sensors.\newline
- If DNA has ``Paired Categoricals'', the domain must involve matching/comparison.\newline
\newline
\textbf{OUTPUT FORMAT:}\newline
Return a JSON list of domains:\newline
\texttt{[}\newline
\hspace*{1em}\texttt{\{}\newline
\hspace*{2em}\texttt{"domain": "Legal Tech",}\newline
\hspace*{2em}\texttt{"scenario": "Contract Comparison",}\newline
\hspace*{2em}\texttt{"justification": "feat\_1/2 are Contract Types, feat\_3/4/5 are Clause Text."}\newline
\hspace*{1em}\texttt{\},}\newline
\hspace*{1em}\texttt{\{}\newline
\hspace*{2em}\texttt{"domain": "E-Commerce",}\newline
\hspace*{2em}\texttt{"scenario": "Product Duplicate Detection",}\newline
\hspace*{2em}\texttt{"justification": "feat\_1/2 are Categories, feat\_3 is Title, feat\_4/5 are Descriptions."}\newline
\hspace*{1em}\texttt{\}}\newline
\texttt{]} \\
\midrule

You are a Data Simulation Engineer. I will provide the ``Structural DNA'' of a dataset.\newline
Your goal is to generate a ``Mutation Config'' that increases the difficulty of the task for an AI Agent.\newline
\newline
\textbf{INPUT DNA:}\newline
\texttt{\{dna\_json\}}\newline
\newline
\textbf{INSTRUCTIONS:}\newline
1. \textbf{Detect Modality:} (Image, Tabular, etc.)\newline
2. \textbf{Select 3 Corresponding Mutagens according to the modality and DNA:} (Blur, Noise, Typos...)\newline
3. \textbf{Output Format:} Return ONLY the JSON Mutation Config (the ``patch'').\newline
\newline
\textbf{EXAMPLE OUTPUT (for an Image Task):}\newline
\texttt{\{}\newline
\hspace*{1em}\texttt{"signal\_mutations": [}\newline
\hspace*{2em}\texttt{\{"type": "salt\_pepper\_noise", "amount": 0.05\},}\newline
\hspace*{2em}\texttt{\{"type": "rotation", "degrees": 45\},}\newline
\hspace*{2em}\texttt{\{"type": "class\_imbalance", "ratio": "1:10"\}}\newline
\hspace*{1em}\texttt{]}\newline
\texttt{\}} \\
\midrule

You are a Synthetic Data Architect. You will receive an \textbf{Abstract Data DNA} (structural skeleton), a \textbf{Target Domain} (semantic context), and a \textbf{Noise Configuration} (difficulty modifiers).\newline
\newline
\textbf{Goal:}\newline
Merge these inputs to generate a \textbf{Concrete Task Specification} (New DNA). You must:\newline
\newline
1. \textbf{Concrete Mapping:} Rename abstract features (e.g., \texttt{feat\_0}) to realistic domain-specific names (e.g., \texttt{systolic\_blood\_pressure}).\newline
2. \textbf{Apply Dimensions:} specify the final row/column counts based on the constraints.\newline
3. \textbf{Embed Logic:} Define the ``Hidden Ground Truth Function'' that relates the features to the target, incorporating the requested noise.\newline
\newline
\textbf{Inputs Provided:}\newline
1. \texttt{\{\{ORIGINAL\_DNA\}\}}: The JSON skeleton from the seed task.\newline
2. \texttt{\{\{SELECTED\_DOMAIN\}\}}: The industry context (e.g., ``Healthcare: Triage'').\newline
3. \texttt{\{\{NOISE\_CONFIG\}\}}: The specific boosters (e.g., ``Starvation Mode + Label Flipping'').\newline
\newline
\textbf{Output Format:}\newline
Return valid JSON only. The structure could be changed due to Tabular, Image, Text, Audio, or Graph data, depending on the original DNA.:\newline
\newline
\texttt{\{}\newline
\hspace*{1em}\texttt{"task\_name": "String (Creative Title)",}\newline
\hspace*{1em}\texttt{"domain\_context": "String",}\newline
\hspace*{1em}\texttt{"final\_dimensions": \{ "n\_samples": Int, "n\_features": Int \},}\newline
\hspace*{1em}\texttt{"feature\_mapping": \{}\newline
\hspace*{2em}\texttt{"feat\_0": \{ "new\_name": "String", "generation\_logic": "String (distribution)" \},}\newline
\hspace*{2em}\texttt{"feat\_1": \{ "new\_name": "String", "generation\_logic": "String" \}}\newline
\hspace*{1em}\texttt{\},}\newline
\hspace*{1em}\texttt{"hidden\_rule\_logic": "String (The mathematical formula: y = f(x) + noise, please do not be so simple, make it complex and realistic)",}\newline
\hspace*{1em}\texttt{"evaluation\_specs": \{}\newline
\hspace*{2em}\texttt{"metric": "String (same as original DNA)",}\newline
\hspace*{2em}\texttt{"thresholds\_logic": "String for Kaggle medal threshold (e.g. Gold = 0.90 (due to 10\% noise). Silver = Random Forest baseline. Bronze = Linear Regression baseline. Median = Majority Class baseline.)"}\newline
\hspace*{1em}\texttt{\}}\newline
\texttt{\}}\newline
\newline
\textbf{Constraints:}\newline
* \textbf{Speed Constraint:} Keep \texttt{n\_samples} to be a random number between 50 and 200.\newline
* \textbf{Consistency:} If the DNA says \texttt{feat\_1} is ``High Cardinality'', map it to something like ``ZipCode'' or ``PatientID'', not ``Gender''.\newline
* \textbf{Logic:} The \texttt{hidden\_rule\_logic} must explicitly use the features you just named. \\
\end{longtable}

\newpage
\begin{longtable}{p{0.95\textwidth}}
\caption{Prompt template for the MLE Developer agent.}
\label{tab:mle_developer_prompt} \\
\toprule
\textbf{MLE Developer Prompt Template} \\
\midrule
\endfirsthead

\multicolumn{1}{c}%
{{\bfseries Table \ref{tab:mle_developer_prompt} continued from previous page}} \\
\toprule
\textbf{MLE Developer Prompt Template} \\
\midrule
\endhead

\midrule
\multicolumn{1}{r}{{Continued on next page}} \\
\endfoot

\bottomrule
\endlastfoot

You are an expert Python MLE Developer. You receive a Synthetic Task Blueprint (JSON) and must write one self-contained Python script named \texttt{generate\_task\_env.py} to build the training environment.\newline
\newline
\textbf{Inputs:}\newline
- Task DNA (JSON): includes modality (Tabular/Image/Text/Audio), \texttt{final\_dimensions}, \texttt{feature\_mapping}, and \texttt{hidden\_rule\_logic}.\newline
\newline
\textbf{What your script must do:}\newline
1) Asset generation (deterministic, no external downloads):\newline
\hspace*{1em}- Tabular: build DataFrames per the blueprint and save to \texttt{train.csv} and \texttt{test.csv} (test $\sim$20\% the size of train).\newline
\hspace*{1em}- Image: create \texttt{images/}, draw synthetic images with Pillow or cv2 following the blueprint, save \texttt{.png}, and create \texttt{train.csv} and \texttt{test.csv} mapping filename to label (test labels blank but column present; schema matches train).\newline
\hspace*{1em}- Audio: create \texttt{audio/}, synthesize waveforms with numpy + scipy.io.wavfile (sine, noise, etc.), save \texttt{.wav}, and create \texttt{train.csv} and \texttt{test.csv} mapping filename to label (test labels blank but column present; schema matches train).\newline
\hspace*{1em}- Text: if short texts, keep directly in \texttt{train.csv}/\texttt{test.csv}; if long documents, save to \texttt{docs/} as \texttt{.txt} and reference from \texttt{train.csv}/\texttt{test.csv} (schema matches).\newline
\newline
2) Hidden rule logic:\newline
\hspace*{1em}- Implement the blueprint's \texttt{hidden\_rule\_logic} to assign labels deterministically. Be explicit about how features drive labels.\newline
\newline
3) Heuristic leaderboard / sanity check:\newline
\hspace*{1em}- Implement the threshold logic described in the blueprint (use its rules, ignore specific numeric targets).\newline
\hspace*{1em}- Compute thresholds on the generated test data (per the threshold logic).\newline
\hspace*{1em}- Save the computed thresholds in \texttt{threshold.json} with exactly these keys: \texttt{"gold\_threshold"}, \texttt{"silver\_threshold"}, \texttt{"bronze\_threshold"}, \texttt{"median\_threshold"} (values derived from the logic).\newline
\newline
4) Sample submission:\newline
\hspace*{1em}- Create \texttt{sample\_submission.csv} with test IDs and a placeholder prediction column matching the target format; fill predictions with random or dummy values as examples.\newline
\hspace*{1em}- Also create \texttt{answer.csv} containing the true labels for test data in the same format as \texttt{sample\_submission.csv} (for hidden evaluation).\newline
\newline
\textbf{Constraints:}\newline
- Script must be standalone and runnable with common Python libs (numpy, pandas, Pillow/cv2, scipy).\newline
- Respect \texttt{final\_dimensions} (n\_samples/n\_features/resolution) from the DNA.\newline
- Use informative comments where needed.\newline
\newline
\textbf{Task DNA:}\newline
\texttt{\{task\_dna\}} \\
\end{longtable}

\newpage
\begin{longtable}{p{0.95\textwidth}}
\caption{Prompt template for the MLOps Engineer agent.}
\label{tab:mlops_engineer_prompt} \\
\toprule
\textbf{MLOps Engineer Prompt Template} \\
\midrule
\endfirsthead

\multicolumn{1}{c}%
{{\bfseries Table \ref{tab:mlops_engineer_prompt} continued from previous page}} \\
\toprule
\textbf{MLOps Engineer Prompt Template} \\
\midrule
\endhead

\midrule
\multicolumn{1}{r}{{Continued on next page}} \\
\endfoot

\bottomrule
\endlastfoot

You are an expert Python MLOps Engineer. Write a robust, standalone evaluation script named \texttt{evaluator.py} for this task.\newline
\newline
\textbf{Task DNA (guidance for metric/dataset logic):}\newline
\texttt{\{task\_dna\}}\newline
\newline
\textbf{Training data generator (use for consistent metric/threshold logic):}\newline
\texttt{\{task\_generator\}}\newline
\newline
\textbf{Inputs to bake into the script:}\newline
- Metric: choose the metric from the Task DNA/training data generator; hardcode its name and direction (\texttt{is\_lower\_better} bool) in the script. Make sure the metric matches the code used in training data generator to derive the thresholds.\newline
- Medal thresholds (JSON): \texttt{\{thresholds\_json\}}\newline
- Submission schema: \texttt{\{schema\}}\newline
\newline
\textbf{Data layout:}\newline
- Public folder contains \texttt{sample\_submission.csv} (user submissions match this schema).\newline
- Ground truth is in \texttt{answer.csv} (relative to \texttt{evaluator.py}).\newline
\newline
\textbf{Script requirements:}\newline
1) Hardcode metric, direction (\texttt{is\_lower\_better} bool), and thresholds at the top (derive metric/direction from Task DNA + schema).\newline
2) CLI: accept \texttt{--submission\_path} (default: \texttt{sample\_submission.csv}).\newline
3) Load submission and ground truth, merge on id column if present; if no id, align by row order with a warning.\newline
4) Compute the specified metric.\newline
5) Output a JSON to stdout with keys: \texttt{"score"}, \texttt{"gold\_threshold"}, \texttt{"silver\_threshold"}, \texttt{"bronze\_threshold"}, \texttt{"median\_threshold"}, \texttt{"is\_lower\_better"}. Use the thresholds as given.\newline
6) On any error, print a JSON error object to stdout (not stderr) and exit gracefully.\newline
7) No external configs or downloads. Use only standard libraries plus numpy/pandas/sklearn if needed.\newline
8) Emit pure Python: no markdown fences, no stray triple quotes. Ensure every string literal is closed and every dict/list is syntactically complete. \\
\end{longtable}

\newpage
\begin{longtable}{p{0.95\textwidth}}
\caption{Prompt template for the Technical Writer agent.}
\label{tab:technical_writer_prompt} \\
\toprule
\textbf{Technical Writer Prompt Template} \\
\midrule
\endfirsthead

\multicolumn{1}{c}%
{{\bfseries Table \ref{tab:technical_writer_prompt} continued from previous page}} \\
\toprule
\textbf{Technical Writer Prompt Template} \\
\midrule
\endhead

\midrule
\multicolumn{1}{r}{{Continued on next page}} \\
\endfoot

\bottomrule
\endlastfoot

You are a technical writer for synthetic ML benchmarks. Using the provided Task DNA, an example description from the seed task, the current contents of the public folder, and the data generator code, write a concise, clear \texttt{description.md} for this synthetic task. Keep structure and tone similar to the example but update names, data details, and task specifics to match the new Task DNA.\newline
\newline
\textbf{Requirements:}\newline
- Include an overview of the problem, data format, and evaluation metric.\newline
- Tell the users to generate the \texttt{submission.csv} file with predictions for the test set, consistent format with the \texttt{sample\_submission.csv}.\newline
- Mention the \texttt{sample\_submission.csv} schema explicitly so users know expected columns.\newline
- List the public data files by name (e.g., \texttt{train.csv}, \texttt{test.csv}, \texttt{sample\_submission.csv}, \texttt{images/*}, \texttt{audio/*}, \texttt{docs/*}) so users know what is provided. Do not include python files or \texttt{answer.csv}.\newline
- Keep the description to have the consistent structure with the example description.\newline
- Do not include Markdown code fences in the output; produce plain Markdown content only.\newline
\newline
\textbf{Task DNA:}\newline
\texttt{\{task\_dna\}}\newline
\newline
\textbf{Example description (seed task):}\newline
\texttt{\{example\_description\}}\newline
\newline
\textbf{Public folder contents:}\newline
\texttt{\{public\_listing\}}\newline
\newline
\textbf{Generator code (for reference on how files are created and what they mean):}\newline
\texttt{\{generator\_code\}}\newline
\newline
\textbf{Sample submission schema:}\newline
\texttt{\{sample\_schema\}} \\
\end{longtable}

\end{document}